\documentclass[journal]{IEEEtran}
\usepackage{amsmath,amsfonts}
\usepackage{algorithm}
\usepackage{array}
\usepackage[caption=false,font=normalsize,labelfont=sf,textfont=sf]{subfig}
\usepackage{textcomp}
\usepackage{stfloats}
\usepackage{url}
\usepackage{verbatim}
\usepackage{graphicx}
\usepackage{cite}
\usepackage{amsmath}
\usepackage{amssymb}
\usepackage{mathtools}
\usepackage{amsthm}
\usepackage{adjustbox}
\usepackage{subfig}
\usepackage{algpseudocode}
\usepackage{enumitem}
\usepackage[symbol]{footmisc}

\usepackage{multirow}
\usepackage{multicol}
\usepackage{wrapfig}
\usepackage[table]{xcolor}
\usepackage[hidelinks]{hyperref}
\usepackage[T1]{fontenc}

\begin{document}

\title{Augmenting Unsupervised Reinforcement Learning with Self-Reference}

\author{
  Andrew Zhao\textsuperscript{\rm 1\textasteriskcentered}\thanks{\textasteriskcentered~Equal contribution.},
  Erle Zhu\textsuperscript{\rm 1\textasteriskcentered},
  Rui Lu\textsuperscript{\rm 1},
  Matthieu Lin\textsuperscript{\rm 2},\\
  Yong-Jin Liu\textsuperscript{\rm 2},~\IEEEmembership{Senior Member,~IEEE}, and
  Gao Huang\textsuperscript{\rm 1\textdagger}\thanks{\textdagger~Corresponding author.},~\IEEEmembership{Member,~IEEE} \\
  \textsuperscript{\rm 1} Department of Automation, BNRist, Tsinghua University, \\
  
  \textsuperscript{\rm 2} Department of Computer Science, BNRist, Tsinghua University \\
  \texttt{\{zqc21,zel20,r-lu21,lyh21\}@mails.tsinghua.edu.cn}, \\
  \texttt{\{liuyongjin,gaohuang\}@tsinghua.edu.cn}
}


\maketitle

\begin{abstract}

Humans possess the ability to draw on past experiences explicitly when learning new tasks and applying them accordingly. We believe this capacity for self-referencing is especially advantageous for reinforcement learning agents in the unsupervised pretrain-then-finetune setting. During pretraining, an agent's past experiences can be explicitly utilized to mitigate the nonstationarity of intrinsic rewards. In the finetuning phase, referencing historical trajectories prevents the unlearning of valuable exploratory behaviors. Motivated by these benefits, we propose the Self-Reference (SR) approach, an add-on module explicitly designed to leverage historical information and enhance agent performance within the pretrain-finetune paradigm. Our approach achieves state-of-the-art results in terms of Interquartile Mean (IQM) performance and Optimality Gap reduction on the Unsupervised Reinforcement Learning Benchmark for model-free methods, recording an 86\% IQM and a 16\% Optimality Gap. Additionally, it improves current algorithms by up to 17\% IQM and reduces the Optimality Gap by 31\%. Beyond performance enhancement, the Self-Reference add-on also increases sample efficiency, a crucial attribute for real-world applications.

\end{abstract}


\section{Introduction}
\label{introduction}
Unsupervised Reinforcement Learning (URL), a paradigm in reinforcement learning (RL), is designed to mimic the effectiveness of the pretrain-finetune framework prevalent in computer vision (CV) and natural language processing (NLP). It involves a two-stage process. The first stage, known as pretraining (PT), focuses on allowing the agent to thoroughly explore and understand its environment and various behaviors. The second stage, finetuning (FT), incorporates extrinsic rewards to guide the agent in addressing a specific downstream task. This is done under the assumption that the environment's transition dynamics are stable. The overarching goal of unsupervised RL is to gather extensive domain knowledge during the pretraining stage, independent of any task-specific reward signals. This knowledge is then efficiently applied in the finetuning stage to tackle downstream tasks. This approach has been explored and supported in various studies \cite{laskin2022cic,zhao2022mixture,pathak2017curiosity,liu2021behavior,liu2021aps,burda2018exploration,pathak2019self,eysenbach2018diversity,yarats2021reinforcement,lee2019efficient}.

\begin{figure*}[ht]
    \centering
        \includegraphics[width=0.98\textwidth]{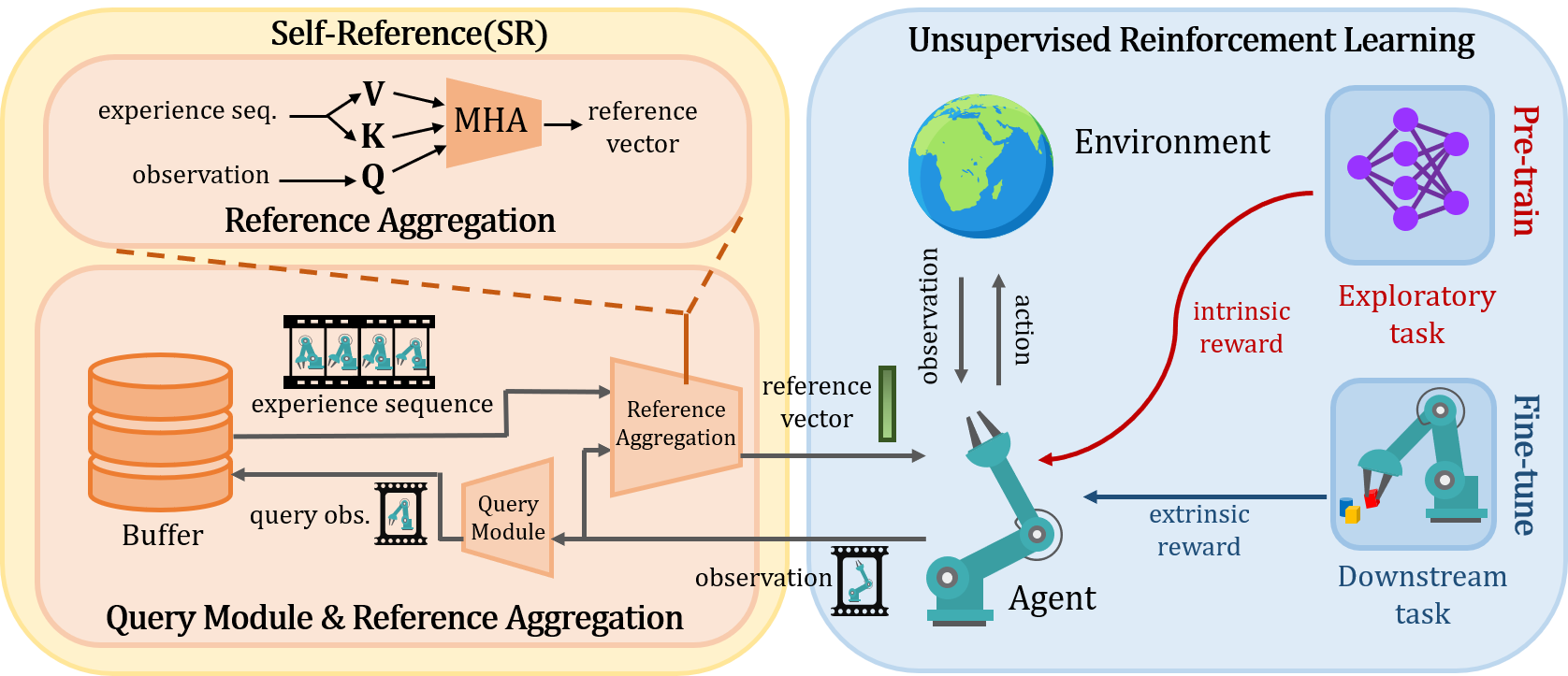}
        \caption{\textbf{Unsupervised Reinforcement Learning with Self-Reference (SR).} The schematic illustration of our method. Given a state, Self-Reference forms a query to the buffer and receives top $k$ neighbors and their subsequent $D$ states. Then, SR aggregates the transitions into a reference vector that aids the agent in learning under the URL setting.}
        \label{fig:Self-Reference}
\end{figure*}

Existing unsupervised reinforcement learning algorithms generally use a measure of surprise as a reward function to explore the environment during the pretraining stage \cite{zhao2022mixture}. As pretraining progresses, this reward function needs to decrease the reward of frequently visited states and give high rewards to less visited states \cite{bellemare2016unifying}. This formulation results in a reward function that implicitly depends on the history of states visited by the agent. However, most current popular URL algorithms do not model this change in reward function \cite{laskin2021urlb}, leading to a nonstationary Markov Decision Process (MDP), which could potentially make learning unstable and inefficient \cite{choi1999environment,sekar2020planning}. Furthermore, works have shown that the naive pretrain-then-finetune paradigm, like the one in unsupervised RL scenarios, can result in unlearning of the pretrained policy's exploratory behaviors in the early finetuning phase \cite{campos2021beyond,wolczyk2023role}. Losing the characteristics of strong exploratory behavior is counterproductive in efficient downstream adaptation scenarios like in the unsupervised reinforcement learning settings. We conjecture that training an agent with awareness of the history of its own transitions could alleviate the problems mentioned.

We propose an add-on method called Self-Reference (SR) for unsupervised reinforcement learning to more effectively utilize historical information during training. This add-on module can improve performance and efficiency during the pretrain and finetune phases of unsupervised reinforcement learning settings. Concretely, we present historical experiences to the agent at every decision epoch. The agent can use these experiences to create summary statistics of visited states during pretraining, explicitly modeling the change in reward. Additionally, by presenting old experiences to the agent, we can also mitigate unlearning of the pretrained exploratory behaviors. Figure \ref{fig:Self-Reference} shows a schematic depiction of Self-Reference. We evaluate SR on the standard Unsupervised Reinforcement Learning Benchmark \cite{laskin2021urlb} and achieve state-of-the-art results for model-free methods by applying Self-Reference to RND \cite{burda2018exploration}. Moreover, experimental results show that our add-on module increases the IQM of APS \cite{liu2021aps} and ProtoRL \cite{yarats2021reinforcement} by as much as 17\% IQM while decreasing Optimality Gap (OG) by 11\% on average and reducing OG by as much as 31\% for ProtoRL. In addition to performance gains, we show that by adding Self-Reference, the agent approaches its asymptotic performance with fewer pretraining steps, indicating that our method also improves sample efficiency. Lastly, we introduce an optional distillation phase after the finetuning phase to distill SR policy to maintain better performance while having no computation overhead over the baseline methods.

\section{Related Works}
\subsection{Retrieval-augmented Techniques in Machine Learning}
Information retrieval is a common task in machine learning. Its solutions recently started to permeate to other subfields to alleviate memorization in parametric weights like in Large Language Model \cite{khandelwal2019generalization,guu2020retrieval,borgeaud2021improving, schick2023toolformer, mialon2023augmented, qin2023tool}, Multi-Modal Modeling \cite{chen2022re}, Decision-making \cite{zhao2023expel}, and Time Series Forecasting \cite{jing2022retrieval}. Moreover, retrieval-augmented systems have been shown to improve neural network performance and generalization capabilities while obtaining more efficient training. In addition to the possible reduction in memorization burden to the network, we apply retrieval to combat nonstationarity and forgetting exploratory behaviors in the pretraining and finetuning phases.

\subsection{Retrieval in Imitation and Reinforcement Learning}
Since traditional imitation learning has exhibited limited scalability due to high data supervision requirements and weak generalization, researchers attempted to take advantage of prior data from previous tasks to adapt models to new tasks more robustly and efficiently \cite{nasiriany2022learning}. As a result, it is found that retrieving prior data and internalizing knowledge from experience achieves considerable progress in novel tasks.

In recent years, retrieval techniques have been brought into the reinforcement learning field. The data buffer and the aggregation mechanism between retrieval data and the model's input are key components in these retrieval systems. The data buffer can be a growing set of the agent's own experiences like the replay buffer \cite{goyal2022retrieval}, or an external dataset from previous experiments \cite{humphreys2022large}, or comes from data collected from other agents and experts as in Offline RL or Imitation Learning \cite{nasiriany2022learning}. The aggregation mechanism for retrieval-augmented RL is to select top-k nearest neighbor sequences in the data buffer and use multi-head attention \cite{vaswani2017attention}, or a specific fusion encoding network \cite{humphreys2022large} 
to aggregate the neighbor sequences and the query states. All the mentioned works focus on querying neighbors based on the encoded observation. We differ from previous work by introducing a more dynamic way of querying the database by training a separate query module. Experiments have shown its strengths of dynamically adapting to different situations over only querying using the current state.

\subsection{Catastrophic Forgetting in Transferring Policy}
Finetuning a pretrained policy is a beneficial approach to transferring knowledge compared to training from scratch. However, recent work discovers that this approach may suffer the problem of catastrophic forgetting \cite{campos2021beyond,wolczyk2023role}, that is, the agent may forget the exploratory skills learned in pretraining since learning signals from the downstream tasks quickly supersede the weights. Behavior Transfer (BT) \cite{campos2021beyond} is proposed as a solution, which utilizes a pretrained policy for exploration, and complements the downstream policy. The result shows that pretrained policies can be leveraged by BT to discover better solutions than without pretraining, and combining BT with standard finetuning results in additional benefits. Our work also focused on addressing this observation and combats this issue by explicitly replaying snapshots of old behaviors to the agent at every decision epoch.

\subsection{Distillation in Reinforcement Learning}
The technique of transferring knowledge from one policy to another, commonly referred to as distillation, holds significant importance in deep reinforcement learning literature. This class of approaches have demonstrated impressive outcomes in enhancing agent optimization, leading to quicker and more robust performance in complex domains \cite{teh2017distral,czarnecki2018mix,gangwani2018policy}. One method is expected entropy regularised distillation \cite{pmlr-v89-czarnecki19a}, allowing quicker learning in various situations. Our work uses the distillation technique to eliminate extra computation during the subsequent deployment stage after training the agent.

\section{Preliminaries and Notations}
\subsection{Markov Decision Process and Reinforcement Learning}
We follow the standard reinforcement learning assumption that the system can be described by a Markov Decision Process (MDP) \cite{sutton2018reinforcement}. An MDP is represented by a tuple $(\mathcal{S}, \mathcal{A}, p, r, \gamma)$, which has states space $\mathcal{S}$, action space $\mathcal{A}$, transition dynamics $p(\mathbf{s}'|\mathbf{s},\mathbf{a})$, reward function $r(\mathbf{s},\mathbf{a},\mathbf{s}')$ and discount factor $\gamma$. At each time step, an agent observes the current state $\mathbf{S} \in \mathcal{S}$, performs an action $\mathbf{A} \in \mathcal{A}$, and then observes a reward and next state $\mathbf{R},\mathbf{S}'\sim r, p$ from the environment. Generally, $r$ refers to an extrinsic reward $r^\text{ext}$, given by the environment. However, in our setting, $r$ could also be an intrinsic reward $r^\text{int}$ generated by the agent using URL algorithms.

An RL algorithm aims to train an agent to interact with the environment and maximize the expected discounted return $G_\pi(\mathbf{s})=\mathbb{E}[\sum^\infty_{t\ge0}\gamma^tR_t]$, where $\pi$ denotes the policy of the agent. Most popular RL algorithms use the action-value function $Q_\pi$ as the optimization objective. It describes the expected return after taking the action $\mathbf{A}_t$ in state $\mathbf{S}_t$ and following parameterized policy $\pi_\theta$ thereafter:
\begin{equation}
Q_{\pi}(\mathbf{s}, \mathbf{a}) = \mathbb{E}_{\pi\theta}[G_t|\mathbf{S}_t=\mathbf{s}, \mathbf{A}_t=\mathbf{a}].
\end{equation}

\subsection{Unsupervised Reinforcement Learning}
Popular deep RL algorithms perform well in many decision-making tasks \cite{sac,lillicrap2015continuous}. However, an agent trained by these algorithms is only capable of the task at hand, with poor generalization ability to transfer to other tasks \cite{kirk2021survey}. URL approaches are designed to train agents with self-supervised rewards with more generalizable policies that can efficiently adapt to downstream tasks \cite{laskin2021urlb}. The Unsupervised Reinforcement Learning Benchmark (URLB) is designed to evaluate the performance of URL algorithms. The benchmark splits the training procedure into two phases: pretraining and finetuning. Agents are trained with intrinsic rewards for $N_{PT}$ steps in the pretraining phase and then evaluated using the performances achieved by finetuning with extrinsic rewards for $N_{FT}$ steps.

\section{Self-Reference}
The unsupervised reinforcement learning (URL) paradigm encompasses two distinct phases: pretraining and finetuning. During the pretraining phase, the reward function's implicit dependence on a history of transitions results in a nonstationary Markov Decision Process, provided the agent does not explicitly account for the evolving reward structure. In Appendix \ref{sec:motivation_pretrain}, we present an illustrative example through an experiment within a multi-armed bandit framework. Here, the intrinsic "moving" reward is represented by counts. We demonstrate that an agent which incorporates historical context into its environmental model incurs lower regret. This translates to more effective learning and exploration during the URL pretraining phase. Inspired by this observation, we suggest the strategic retrieval of relevant historical data to address the challenges posed by nonstationarity.

Furthermore, works like \cite{humphreys2022large} have shown that referencing expert demonstrations benefits learning their behavior. In the unsupervised reinforcement learning setting during finetuning, we can view some exploratory behaviors as beneficial for quickly learning the new task and as good reference trajectories for the agent to maintain some of its favorable characteristics. Therefore, retrieving relevant good exploratory behaviors should also be beneficial for URL during finetuning.

Equipped with the intuition that explicitly using past experiences as references for the agent could benefit the agent in the URL setting both during pretrain and finetune, we devise a new module for unsupervised reinforcement learning algorithms called Self-Reference (SR). We designed this module as an add-on method that can boost any unsupervised reinforcement learning method's performance. The Self-Reference module consists of a query module and a network aggregating the queried experiences as additional knowledge. Readers can find a schematic figure of Self-Reference in Figure \ref{fig:Self-Reference}.

\subsection{The Query Module}
\label{sec:method_query_module}
In order to query informative trajectories from historical experiences, we need to develop a module that can retrieve appropriate values that could be useful for the unsupervised reinforcement learning agent. We argue that the query module should have the flexibility to query the following trajectories:
\begin{itemize}[noitemsep, leftmargin=*]
    \item  \textit{Neighbors.} Intuitively, neighbors can show where the agent has been in close proximity and infer where to explore more.
    \item \textit{Partial neighbors.} In some cases, we only need to select reference neighbors with particular features and explore different parts of the feature space.
    \item \textit{Information on possible future visitations.} Since the agent needs to maximize the state-value function, it needs to be forward-seeking and have the option to look at other states that the agent might wish to visit.
\end{itemize}
To enhance the system's adaptability, we developed a module capable of determining the optimal queries. This module, denoted as $\pi^\text{query}_{\phi}:\mathcal{S}\rightarrow \mathcal{S}$, takes the current state as input and queries historical data to support the URL agent's decision-making process during training. For optimization, we employed the well-known on-policy algorithm PPO \cite{schulman2017proximal}, aiming to maximize the task return. This encompasses the intrinsic reward during the pretraining phase and the task reward throughout the finetuning stage. Our approach keeps the hyperparameters consistent with those specified in the CleanRL framework \cite{huang2022cleanrl}. To further refine our system, based on the notion that querying for nearest neighbors can serve as a beneficial inductive bias \cite{humphreys2022large,goyal2022retrieval}, we introduced an identity loss $\mathcal{L}_\text{identity}=||\mathbb{E}_{\mathbf{q}\sim\pi_\text{query}(\mathbf{s})}[\mathbf{q}]-\mathbf{s}||_2$ to the query actor, where  $\mathbf{q}\sim\pi_\text{query}\in\mathcal{S}$ represents the generated query. The total loss for the query actor is
\begin{align}
\mathcal{L}_{\pi_\text{query}}=\mathcal{L}_\text{PPO}+\mathcal{L}_\text{identity}.\label{eq:query-agent}
\end{align}
To further facilitate this inductive bias, we do not use the query module for half of the episodes during the PT phase and use the current state $s_t$ as the query. We do not update the query module's parameters on the trajectories that simply use $s_t$ as the query. The value function of the query module, $V^\text{query}$ is trained using the standard PPO value loss.

After obtaining the query, we use \texttt{Faiss} \cite{johnson2019billion} to find the top $k$ nearest neighbors (keys) of that query and retrieve all subsequent states within $D$ timesteps. We hypothesize that expanding the retrieved states to trajectories provides more information like rolling out in planning algorithms (See Section \ref{subsec:hyper} for confirmation). From this retrieval process, we obtain a set $E$ defined as
\begin{align}
E = \{(\mathbf{s}_{t_1}, ... , \mathbf{s}_{t_1+D-1}), ... , (\mathbf{s}_{t_k}, ... , \mathbf{s}_{t_k+D-1})\},\label{eq:retrievals}
\end{align}
which contains $k$ ordered lists, each containing $D$ subsequent states of respective $k$ key states. The subscript under $t$ indicates the $k$th neighbor for the query. With this query module trained, we obtained a viable way to query the experiences for additional knowledge under appropriate situations. We will leave retrieving the action and reward from historical trajectories as future work.

\subsection{The Retrieval Mechanism}
\label{sec:method_retrieval}
Since Self-Reference agents need to retrieve historical states, we must have a component that maintains historical information. Fortunately, off-policy reinforcement learning algorithms often already possess a large experience replay buffer $\mathcal{D}$ \cite{lin1992self} in order to train the RL agent; therefore, we can query from this buffer without additional memory constraints. As the agent needs to query historical information exhaustively at every step, we propose using a subset of the experience replay buffer \cite{lin1992self} as retrievable historical experiences for SR agents: $\mathcal{D}^\text{reference}\subset\mathcal{D}$. In all of our experiments, the number of retrievable states is set to the $|\mathcal{D}^\text{reference}|=1e5$ latest transitions unless stated otherwise ($|\mathcal{D}|=1e6$ for comparison). We keep the context window $|\mathcal{D}^\text{reference}|$ capped at the agent's last episode's experiences, avoiding the possibility of the agent querying a state from the current episode.

We utilized \texttt{Faiss} \cite{johnson2019billion} as a scalable method to efficiently employ GPUs when performing similarity searches. We maintained the key-query space identical to the state space because it already contains all the information needed for decision-making. Moreover, cosine similarity was used as the metric when performing a k-NN search, as we found it slightly outperformed $l$-2 norm in our experiments. Please refer to Appendix \ref{sec:faiss} for a detailed description of \texttt{Faiss} and nearest neighbor search.

\subsection{Aggregating Retrieved Experiences}
\label{sec:method_combining}

After retrieving experiences from the replay buffer, we must meaningfully combine them into a feature vector for the critic and actor networks. The attention mechanism has been a popular choice for dynamically combining information \cite{vaswani2017attention}. We employ the multiheaded cross-attention mechanism, setting the current state hidden feature as the query (note that this query is used in the multiheaded attention mechanism but not the one used to query the replay buffer), and features extracted from experiences $E$ (Equation \ref{eq:retrievals}) from the replay buffer as keys and values. We refer readers to the original paper \cite{vaswani2017attention} for more details. Additionally, we incorporate learnable time step embeddings into the keys and values features to indicate to the agent that each transition of $D$ steps has temporal meaning, which will receive gradient updates from the DDPG objective. We refer to the final output of the multiheaded cross-attention module as a \textit{reference vector}, containing additional historical information to aid the agent during training. The procedure for computing the reference vector is as follows ($U$ is the embedding dimension of MHA and the dimension of the reference vector):
\begin{align}
\intertext{Query of MHA:}
Q &= \operatorname{QEncoder}(s_t), \nonumber\\ \quad \text{where } & s_t \text{ is the current state}. \nonumber \\
\intertext{Key of MHA:}
K &= \operatorname{KEncoder}(s_{t_1},\dots,s_{t_{1+D-1}},\ldots, s_{t_k},\dots,s_{t_{k+D-1}}), \nonumber \\ \quad \text{where } & s_{t_j:t_{j+D-1}}\in E. \nonumber \\
\intertext{Value of MHA:}
V &= \operatorname{VEncoder}(s_{t_1},\dots,s_{t_{1+D-1}},\ldots, s_{t_k},\dots,s_{t_{k+D-1}}), \nonumber \\ \quad \text{where } & s_{t_j:t_{j+D-1}}\in E. \nonumber \\
\intertext{Time Embedding:}
T &= \operatorname{TimeEmbed}(t_1, \dots,t_{1+D-1},\ldots, t_k, t_{k+D-1}). \nonumber \\
\intertext{Reference Vector:}
u_t &= \operatorname{MultiHeadAttention}(Q, K+T, V+T) \in \mathbb{R}^{U}.
\label{eq:rf_vec}
\end{align}

Finally, the reference vector is concatenated with the actor network's state feature and the critic network's state and action feature. With the added computation of the SR module, we noticed a mean training Frames-Per-Second (FPS) decrease from 17 to 15, which is relatively marginal due to efficient implementations. We present the pseudo-code for Self-Reference in Algorithm \ref{alg:Self-Reference}.

\subsection{Distillation (Optional)}
\label{sec:method_distillation}
Since Self-Reference introduces additional computation (retrieval and aggregation of retrieved experiences) when the policy performs actions, it might not be a favorable trait for deployment. We observe that since SR is designed to aid the agent's training stage, we can effectively \textit{eliminate the computation overhead} Self-Reference introduces during deployment by performing policy distillation. Specifically, we use the policy after the finetuning phase as a teacher network and a network without the query module, MHA, and retrieval parts as a student network (same network as the baseline methods). After finetuning, we relabel every transition's action in the replay buffer as the teacher network's output and train the student network using maximum log-likelihood, $\mathcal{L}_\text{student}=\mathbb{E}_{\mathbf{s}\sim\mathcal{D},\mathbf{a}\sim\pi_\text{teacher}(\mathbf{s})}[-\log\pi_\text{student}(\mathbf{a}|\mathbf{s})]$. We trained the student policy with the same network as the baselines and employed a cosine learning rate schedule with a learning rate of 1e-3 for 200 epochs.

\begin{figure*}[!htb]
    \centering
    \renewcommand{\thesubfigure}{}
    \subfloat{
        \includegraphics[width=0.98\textwidth]{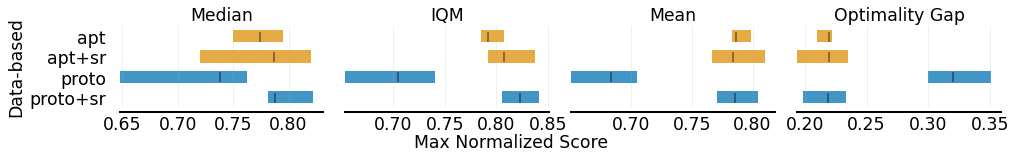}
        \label{fig:main_data}
    }
    \hfill
    \subfloat{
        \includegraphics[width=0.98\textwidth]{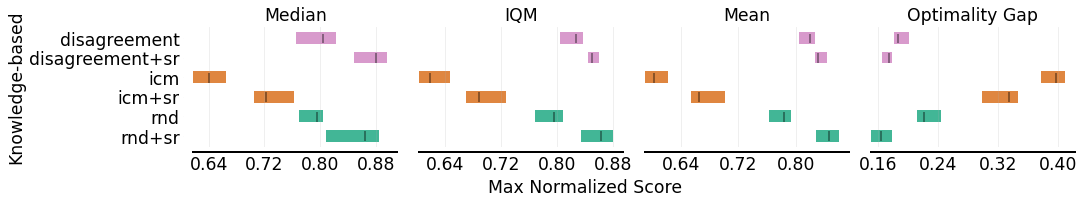}
        \label{fig:main_knowledge}
    }
    \hfill
    \subfloat{
        \includegraphics[width=0.98\textwidth]{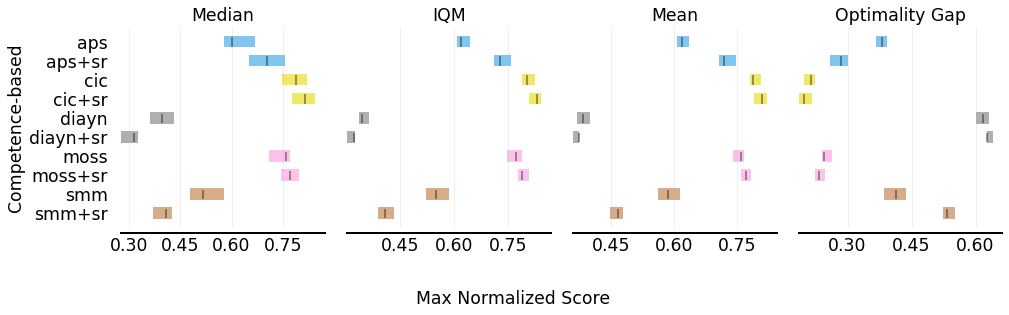}
        \label{fig:main_competence}
    }
    \caption{\textbf{Main Results of Self-Reference on Unsupervised Reinforcement Learning Benchmark.} We showcase our main results, reported using RLiable with statistically robust metrics. Our main results emphasize on the IQM and Optimality Gap metrics and provide Median and Mean for reference purposes.}
    \label{fig:main_results}
\end{figure*}

\begin{figure*}[ht]
\centering
\begin{tabular}{cc}
\subfloat[25K Steps]{\includegraphics[width=0.48\textwidth]{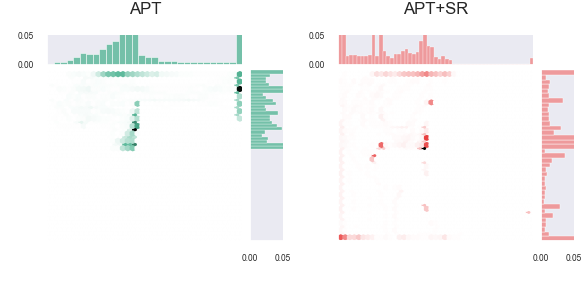}} &
\subfloat[50K Steps]{\includegraphics[width=0.48\textwidth]{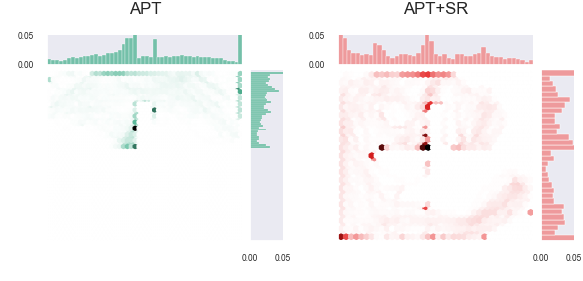}} \\
\subfloat[75K Steps]{\includegraphics[width=0.48\textwidth]{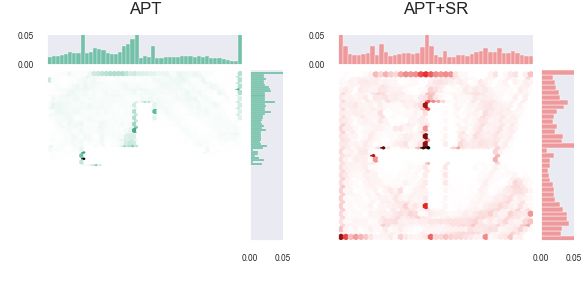}} &
\subfloat[100K Steps]{\includegraphics[width=0.48\textwidth]{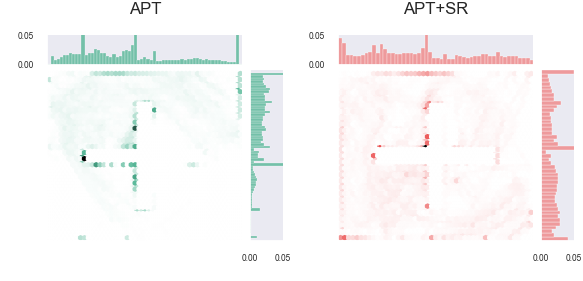}}
\end{tabular}
\caption{\textbf{Visitation Jointplot.} A plot of visitations as density points as PT steps increase, with margin histograms labeled with normalized probability. APT+SR (red) was able to cover the Y-axis in 25k steps and the whole X-Y plane in only 50k steps, while Vanilla APT (green) lags behind and only starts to cover the X-Y plane at 100k PT steps. Note that the empty cross-shape in the middle is formed due to an impenetrable wall that the agent cannot pass through.}
\label{fig:visitation}
\end{figure*}

\section{Experiments}
\subsection{Benchmark and Evaluation}
We apply our method to existing unsupervised reinforcement learning algorithms and evaluate these new methods on tasks from the URL Benchmark (URLB) \cite{laskin2021urlb}, a standard in URL research. Please refer to Appendix \ref{sec:benchmark} for more details on this benchmark and the baselines used.

\textbf{Evaluation:} For our main results, we follow the \textit{URLB's} training procedure by pre-training the agent for two million steps in each domain with intrinsic rewards and finetuning the pre-trained agent for one hundred thousand steps with downstream task rewards. All baseline experiments were conducted for eight seeds \textit{(0-7)} per downstream task for each algorithm using the code from the URL Benchmark. Furthermore, \textit{SR} applied to vanilla URL algorithms are also evaluated for eight seeds per task. A total of $1920 = 2$ (use SR or not) $\times 10$ algorithms $\times 12$ tasks $\times 8$ seeds experiments were conducted for the main results. All methods use the DDPG agent as their backbone.

We use the scores of a DDPG agent trained from scratch for two million steps as the expert scores and normalize the original task scores. To make the main results more convincing, we use statistical-based metrics \textit{interquartile mean (IQM)} and \textit{Optimality Gap (OG)} of normalized scores as our main evaluation criteria, with mean values and median values as references. \textit{IQM} is more unbiased than the median, while \textit{Optimality Gap} is the distance between a method's score and expert scores \cite{agarwal2021deep}. We present all results with error bars using standard error.

\begin{table*}[ht]
\caption{\textbf{Numerical Results.} The numerical results of our approach compared with baselines. Note values are mean scores of all the tasks ± the standard error across eight seeds, which might be a less robust result than IQM. Presented for ease of comparison to previous works.}
\centering
\resizebox{\textwidth}{!}{%
\begin{tabular}{l|llll|llll|llll}
\hline
Domain & \multicolumn{4}{c|}{Walker} & \multicolumn{4}{c|}{Quadruped} & \multicolumn{4}{c}{Jaco Reach} \\
Task & Flip & Run & Stand & Walk & Jump & Run & Stand & Walk & Bottom Left & Bottom Right & Top Left & Top Right \\
\hline
 \texttt{\textbf{\textcolor{blue}{Baselines}}} & & & & & & & & & & & & \\
\hline
\rowcolor{gray!25}APS & 535±27 & 314±29 & 843±28 & 808±66 & 490±30 & 378±25 & 771±35 & 494±53 & 124±11 & 104±14 & 120±10 & 110±12 \\
APT & 682±28 & 482±21 & 935±16 & 925±9 & 693±22 & 463±15 & 890±19 & 831±52 & 131±5 & 141±7 & 146±5 & 155±4 \\
\rowcolor{gray!25} CIC & 639±37 & 428±26 & 952±6 & 873±26 & 713±26 & 500±6 & 917±3 & 886±12 & 149±6 & 147±7 & 140±7 & 143±6 \\
DIAYN & 330±16 & 208±9 & 845±29 & 608±44 & 537±33 & 338±24 & 776±27 & 367±19 & 3±0 & 6±1 & 15±3 & 9±2 \\
\rowcolor{gray!25} Disagreement & 610±22 & 474±26 & 960±3 & 913±15 & 756±8 & 571±15 & 934±13 & 896±8 & 142±5 & 167±5 & 149±4 & 148±4 \\
ICM & 502±17 & 241±24 & 927±13 & 757±24 & 423±40 & 271±18 & 529±54 & 262±14 & 146±6 & 133±4 & 161±4 & 145±10 \\
\rowcolor{gray!25}MOSS & 609±27 & 452±19 & 964±1 & 879±12 & 711±32 & 473±22 & 922±10 & 757±59 & 145±9 & 139±8 & 114±9 & 144±7 \\
Proto & 511±19 & 239±10 & 922±21 & 895±11 & 511±45 & 348±31 & 692±65 & 556±58 & 151±10 & 151±7 & 147±13 & 162±7 \\
\rowcolor{gray!25} RND & 698±23 & 369±19 & 946±13 & 854±29 & 756±7 & 521±7 & 921±9 & \textbf{905±4} & 142±3 & 142±5 & 136±4 & 128±7 \\
SMM & 549±32 & 247±23 & 943±4 & 769±57 & 538±45 & 437±19 & 792±41 & 466±52 & 85±6 & 94±7 & 83±6 & 97±8 \\
\hline
\texttt{\textbf{\textcolor{orange}{Ours}}} & & & & & & & & & & & & \\
\hline
\rowcolor{gray!25} APS+SR & 734±36 & 478±36 & 938±13 & \textbf{944±6} & 560±39 & 364±25 & 717±41 & 601±76 & 145±5 & 144±5 & 115±6 & 133±6 \\
APT+SR & 692±28 & \textbf{555±21} & 969±4 & 939±6 & 638±48 & 426±22 & 811±42 & 689±70 & 133±7 & 166±6 & 148±4 & 160±9 \\
\rowcolor{gray!25} CIC+SR & 706±31 & 501±23 & 944±13 & 890±25 & \textbf{794±11} & 546±12 & 930±5 & 904±7 & 138±8 & 120±6 & 142±7 & 156±9 \\
DIAYN+SR & 406±8 & 181±9 & 928±12 & 641±12 & 400±55 & 296±39 & 529±26 & 258±30 & 24±3 & 19±3 & 27±3 & 26±3 \\
\rowcolor{gray!25} Disagreement+SR & 718±29 & 444±20 & 947±13 & 919±17 & 769±17 & 546±13 & 869±37 & 839±15 & 140±4 & 165±3 & 170±7 & 172±4 \\
ICM+SR & 535±25 & 343±24 & 887±54 & 710±56 & 394±34 & 309±25 & 655±52 & 298±34 & 163±5 & \textbf{170±9} & \textbf{172±3} & \textbf{181±5} \\
\rowcolor{gray!25} MOSS+SR & 636±35 & 444±32 & 970±1 & 875±23 & 726±32 & 516±38 & 937±6 & 881±26 & 117±8 & 140±5 & 142±9 & 122±7 \\
Proto+SR & \textbf{756±26} & 480±26 & \textbf{972±1} & 932±6 & 555±49 & 439±25 & 826±31 & 663±73 & 154±11 & 155±6 & 151±8 & 173±5 \\
\rowcolor{gray!25} RND+SR & 717±29 & 476±23 & 955±10 & 832±40 & 775±11 & \textbf{608±18} & \textbf{950±4} & 835±42 & \textbf{170±5} & 161±8 & 157±7 & 170±6 \\
SMM+SR & 681±33 & 361±21 & 969±2 & 804±43 & 411±22 & 321±21 & 553±35 & 279±29 & 30±3 & 38±4 & 43±7 & 36±4 \\
\hline
\end{tabular}
}
\label{tab:results}
\end{table*}

\subsection{Main Results}
The main results in URLB are presented in Figure \ref{fig:main_results}; we separate the URL algorithms into three categories and present their before and after performance using SR. All URL algorithms have obtained a 5\% improvement in \textit{IQM} and an 11\% reduction in \textit{OG} on average. To our surprise, APS+SR achieves a 17\% higher \textit{IQM} than vanilla APS, and ProtoRL+SR obtains a 31\% lower \textit{OG} than vanilla ProtoRL. We found that after using SR, data-based methods achieved an 8.5\% higher \textit{IQM} and a 19\% lower \textit{OG} than before on average, representing the best progress among the three categories. We did notice that DIAYN and SMM dropped in performance after adding Self-Reference. We hypothesize that since DIAYN suffers from a lack of exploration \cite{strouse2021learning}, and SR tends to accelerate convergence due to alleviating the nonstationarity problem, in the case of DIAYN, SR could impede exploration even more, resulting in worse performance.

\begin{figure}[ht]
    \begin{center}
        \centerline{\includegraphics[width=\columnwidth]{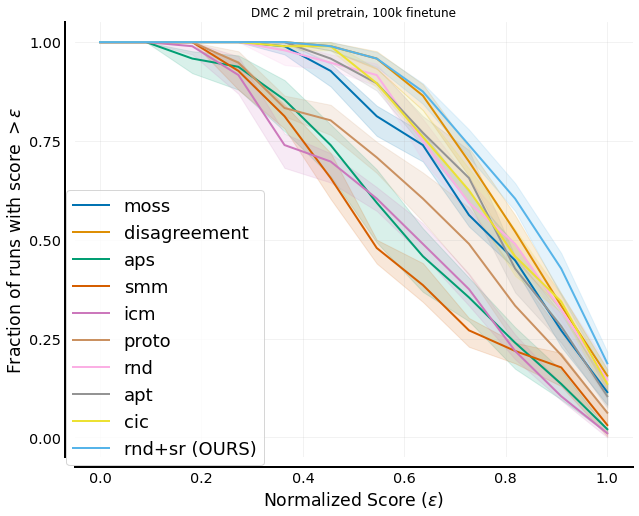}}
        \caption{\textbf{Performance Profile on URL Benchmark.} Performance profile of RND+SR (ours) vs. existing methods. RND+SR stochastically dominates all other methods and obtains SOTA on the URL Benchmark for model-free methods.}
        \label{fig:performance_profile}
    \end{center}
\end{figure}

Figure \ref{fig:performance_profile} shows the performance profile on the URL Benchmark, where the point on the curve reports the probability of achieving the score through the method. RND+SR (Ours) stochastically dominates all other approaches and obtains state-of-the-art (SOTA) performance on URLB.

We also show the numerical results of every baseline method with and without Self-Reference in Table \ref{tab:results}.

\subsection{Pretrain Phase Results}
\label{sec:pretrain_results}
To better understand how our method behaves differently in the pretraining phase alone, we conducted experiments in the point mass maze environment to demonstrate that obtaining information from the history of transitions helps agents learn more efficiently. The point mass maze environment lets the agent control a ball in a 2D plane with a cross-shaped wall in the center where the ball cannot go through. We choose the APT method in this experiment since the intrinsic reward is the entropy of the state visitation, which is greatest when the state visitation is uniformly distributed and can be more intuitively visualized. We train APT and APT+SR for $100$k steps and plot the visitation joint-plot in Figure \ref{fig:visitation}. The plot shows that the APT agent struggles to get out of the left top quadrant pre-25k steps while APT+SR already covers the Y-axis evenly. Then, at 50k steps, the APT+SR agent covers the X-Y plane while the APT agent struggles to visit the lower half of the plane. Finally, at 100k steps, APT+SR achieves a relatively uniform coverage while vanilla APT only started exploring the lower half of the plane. Overall, this result demonstrates that the APT+SR agent covers the X-Y plane much faster than vanilla APT, demonstrating the Self-Reference module's efficacy in learning from the nonstationary task reward of changing state-space coverage, similar to our observation from the MAB experiment in Appendix \ref{sec:motivation_pretrain}.

\subsection{Distillation Phase Results}
We perform the distillation procedure described in Section \ref{sec:method_distillation} on the new SOTA approach, RND+SR, with three seeds on all twelve tasks on URLB. We used the percentage of performance measured against the teacher policy as a metric and obtained $100.1\pm0.3\%$ for Walker, $99.8\pm0.5\%$ for Quadruped, and $96.5\pm1.5\%$ for Jaco. Across three domains $\times$ three seeds $\times$ four tasks, the student agent achieved little to \textbf{no performance degradation}, demonstrating that this simple distillation method \textit{effectively addresses the inference overhead of SR}. We leave more sophisticated distillation methods (e.g., \cite{pmlr-v89-czarnecki19a}) as future work.

\section{Ablation and Empirical Analysis}
\label{ablation}
\begin{figure*}[ht]
\centering
\begin{minipage}[t]{0.48\linewidth}
\centering
\includegraphics[width=\linewidth]{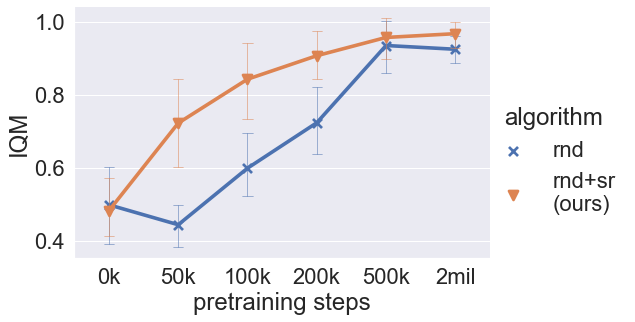}
\caption{\textbf{Pretraining steps vs. performance.} To illustrate the efficiency of our method, we pretrain agents with fewer steps and plot the FT performances in the Quadruped domain. We observe that RND+SR quickly rises in performance while vanilla RND struggles to perform high in fewer pretraining steps scenarios.}
\label{fig:efficiency_plot}
\end{minipage}\hfill
\begin{minipage}[t]{0.48\linewidth}
\centering
\includegraphics[width=\linewidth]{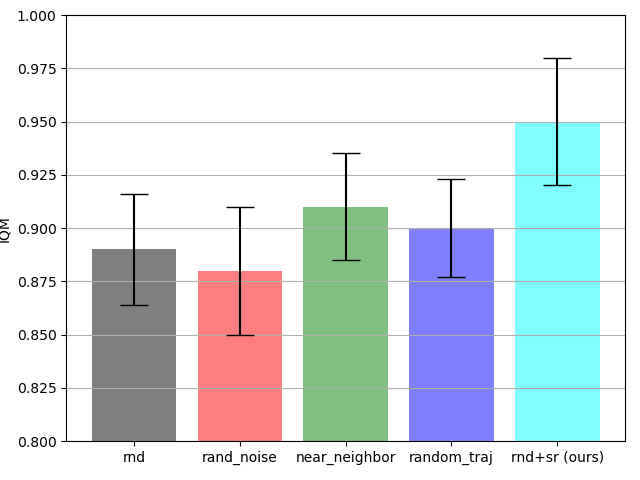}
\caption{\textbf{Ablation on Reference Vector.} We compare SR's query module to different hand-crafted ways of constructing the reference vector. Noticeably, if we only retrieve neighbors based on the current state or randomly, it performs worse than with our query module on the downstream task.}
\label{fig:reference_vector_ablation}
\end{minipage}
\end{figure*}

\textbf{PT-FT Sample Efficiency.} First, we demonstrate how our method can boost sample efficiency under the PT/FT framework. Since RND+SR achieved SOTA performance on URLB, we will use RND as the intrinsic reward and Quadruped as the test domain for the remainder of this Section \ref{ablation} with three seeds to account for variability. For this experiment, we constrain the pretraining steps RND and RND+SR to fewer steps and show the FT performance in Figure \ref{fig:efficiency_plot}. The plot at 0k steps shows DDPG vs. DDPG with Self-Reference, essentially relegating to a supervised RL scenario. It is evident that training from scratch (pretrain steps = 0k) with the Self-Reference module yields no significant gains. This suggests that the incorporation of a retrieval mechanism alone does not substantially aid in the supervised reinforcement learning scenario. However, the development of a robust exploratory policy from PT, coupled with the prevention of forgetting, is essential for achieving performance improvements. Furthermore, in the 50k and 100k pretrain steps scenarios, vanilla RND increased its performance slowly (even decreased in performance with 50k PT steps). At the same time, RND with Self-Reference quickly achieved a high score across Quadruped tasks with a normalized IQM of 0.83 in just 100k pretraining steps. We believe the increase in sample efficiency with the Self-Reference module stems from mitigating the nonstationary reward problem, which makes the agent explore the state space of the environment more efficiently, as shown in Section \ref{sec:pretrain_results} and in our MAB experiment in Figure \ref{mab-regret}.

\textbf{Efficacy and Analysis of Query Module.} We additionally demonstrate how different types of reference vectors affect agent performance and the effectiveness of our query module. Besides our method of using a query actor to select references, we devised three \textit{hand-crafted} approaches for selecting references: 1) combining features from the nearest neighbors of the current state $s_t$ and their subsequent $D$ states, 2) combining features from uniformly sampled states and their subsequent $D$ states, and 3) setting the reference vector $\sim\mathcal{N}(0, 1)$. We present the aggregated results for the Quadruped domain in Figure \ref{fig:reference_vector_ablation}. Utilizing the nearest neighbor of the current state or retrieving random trajectories proves somewhat helpful, illustrating how explicitly providing the agent with the history of transitions is beneficial. Moreover, concatenating random noise to the agent features, essentially denoising does not significantly impact performance. Overall, SR achieves the best result, emphasizing the query module's flexibility, which can either mimic the nearest neighbor queries by learning an identity function, attempt to match random sampling by outputting high entropy queries or query the replay buffer in any other manner beneficial for enhancing the agent's performance.

To further investigate the query module's functionality, we visualize the query module outputs during PT and FT phases in point mass maze environment. Current observations (i.e., current state), query observations (i.e., outputs of query module), and neighbor trajectories (sampled from the buffer) are plotted in blue, red, and green, respectively. To better visualize the agent's behaviors, we increased the color gradient of the state/query as a linear function of time.

\begin{figure}[ht]
    \centering
    \subfloat[Pretrain phase \label{fig:vis_query_pt}]{
        \includegraphics[width=0.48\columnwidth]{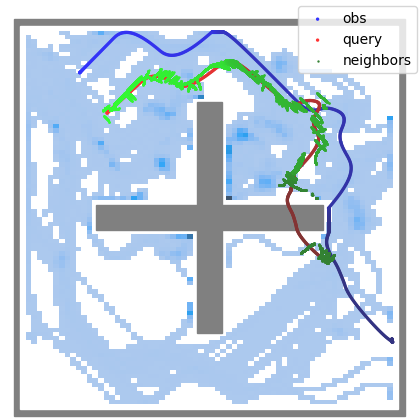}
    }
    \subfloat[Finetune phase \label{fig:vis_query_ft}]{
        \includegraphics[width=0.48\columnwidth]{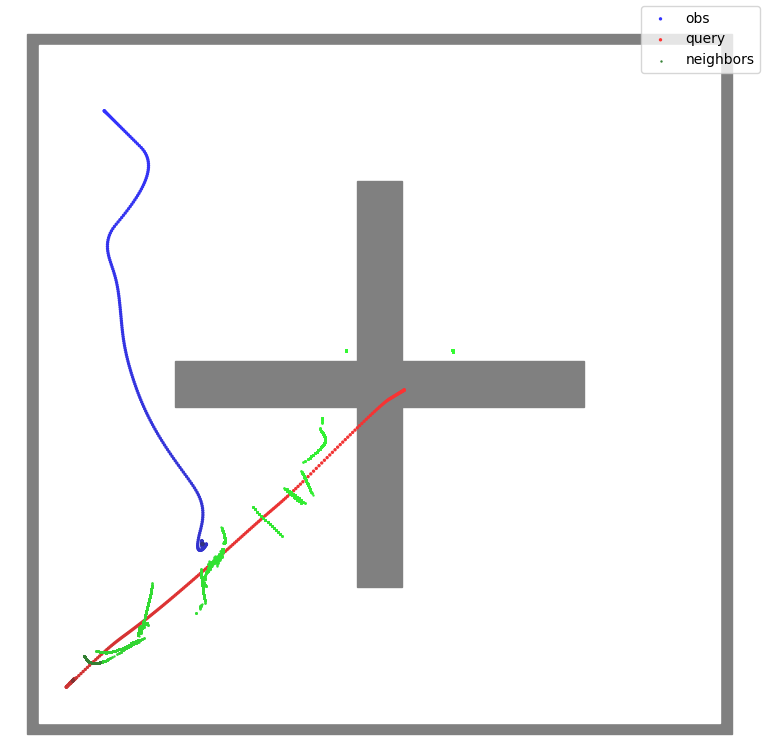}
    }
    \caption{\textbf{Visualization of Query Module in Point Mass Env.} (a) query module's output during PT: the blue background represents visited states, and the query output helps the agent to explore. (b) query module's output during finetuning: query outputs guide the agent to the target position.}
    \label{fig:vis_query}
\end{figure}

In the PT phase, since the agent needs to explore the space, we plot the density of all retrievable states as the intensity of the blue background color. We find that during pretraining, the query module tends to learn to guide the agent's future occupancy and lets it avoid local highly dense areas created by nearest neighbors in the buffer. In Figure \ref{fig:vis_query_pt}, the agent wishes to go to less dense areas to collect more rewards, e.g., the right bottom. The query module creates a "boundary" and guides the main agent to avoid densely populated areas, e.g., the right corner near the cross-shaped wall. For the FT case, we observe that the query module tends to create a "target" where the agent tends to somewhat go towards the queried states. Figure \ref{fig:vis_query_ft} shows the reach bottom left task where the query module learns to create a "destination of point(s)."

\begin{figure*}[ht]
    \centering
    \includegraphics[width=0.98\textwidth]{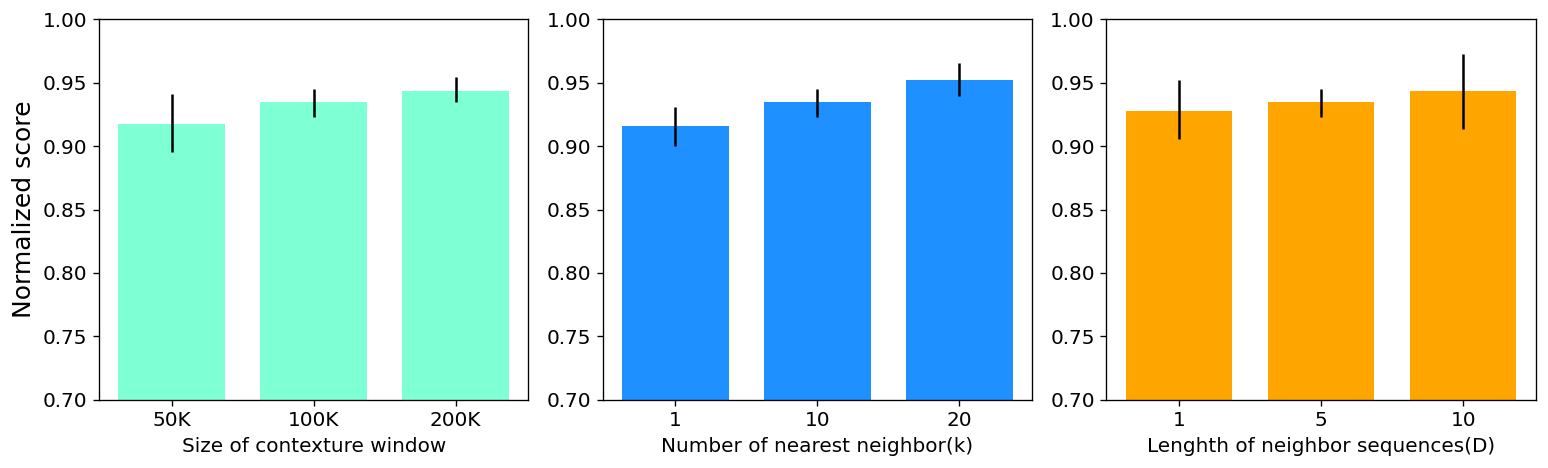}
    \caption{\textbf{Experiments of representative hyperparameters of Self-Reference.}   We use RND+SR and quadruped to evaluate the relationship between the agent's performance and three representative hyperparameters of SR: The number of nearest neighbors $k$, the length of neighbor sequences $D$, and the context window of retrievable states. We noticed a general trend of increasing reference capacity also increases performance.}
    \label{fig:hparam_sel}
\end{figure*}

\textbf{Scaling Hyperparameters Ablation.}
\label{subsec:hyper}
In our methodology, we have identified three critical hyperparameters for SR: the number of nearest neighbors (\(k\)), the length of neighbor sequences (\(D\)), and the context window size for retrievable states. These are set at \(10\), \(5\), and \(100\)k, respectively. To understand how these hyperparameters affect the agent's performance, we conducted experiments using RND+SR with varying values. The experimental results, as illustrated in Figure \ref{fig:hparam_sel}, reveal that an increase in these hyperparameters leads to higher scores for the agent, alongside a greater acquisition of reference information. This suggests that scaling up is beneficial for our method. However, it's important to note that increasing the context window size, the number of nearest neighbors, and the length of neighbor sequences all require additional computational resources. Thus, an optimal balance between computational demands and performance efficiency was established for all three pivotal hyperparameters of SR.

\begin{table}[t]
    \caption{\textbf{Policy Change During Finetuning.} We compare the policy change during finetuning at 10k training steps to the policy at the end of pretraining in the Quadruped domain.}
    \label{table:kl-policy}
    \begin{center}
    \begin{small}
    \begin{sc}
    \begin{adjustbox}{width=\columnwidth}
    \begin{tabular}{|l|c|c|r|}
    \hline
    Task & RND (KL/Norm. Intr.) & RDN + SR (KL/Norm. Intr.) \\
    \hline
    \rowcolor{gray!25} Walk    & 18.9/0.46 & 16.9/0.59\\
    Jump    & 17.5/0.55 & 17.1/0.66\\
    \rowcolor{gray!25} Run     & 18.9/0.59 & 15.9/0.64\\
    Stand   & 18.3/0.74 & 18.3/0.68\\
    \hline
    \end{tabular}
    \end{adjustbox}
    \end{sc}
    \end{small}
    \end{center}
\end{table}

\textbf{Alleviation of Unlearning of PT Policy.} Works such as \cite{campos2021beyond,wolczyk2023role} argue that naively finetuning an agent can exhibit varying degrees of catastrophic forgetting. Since the PT agent behaviors are often purely exploratory, quickly forgetting these exploratory properties might decrease the efficiency during FT in finding the optimal policy. As we explicitly show the agent its old behavior, we hypothesize that this explicit behavior could alleviate the unlearning of helpful exploratory behaviors. To determine if this is the case, we measure the extent of change from the PT behavior to the FT behavior by computing the distance of policy output distributions at every step. Specifically, inspired by KL-control \cite{stengel1986stochastic}, we use the KL divergence from the policy during FT compared to its frozen PT policy as a proxy for the similarity between policies. Furthermore, we also use the drop in intrinsice reward of the FT policy as a proxy for how much the FT policy forgot. Since all FT policies experience the most significant drop in performance of the intrinsic reward at the 10k step, we report the normalized intrinsic return at this step with the average PT intrinsic return as the normalizer. Table \ref{table:kl-policy} showcases the average KL divergence over finetuning and the normalized intrinsic return at 10k steps. We observe that RND + SR found "closer solutions" to the strong exploratory PT policy, confirming our intuition that explicitly showing old behaviors to the agent resulted in less unlearning of the PT policy.

\begin{figure*}[!htb]
    \centering
    \renewcommand{\thesubfigure}{}
    \subfloat{
        \includegraphics[width=0.98\textwidth]{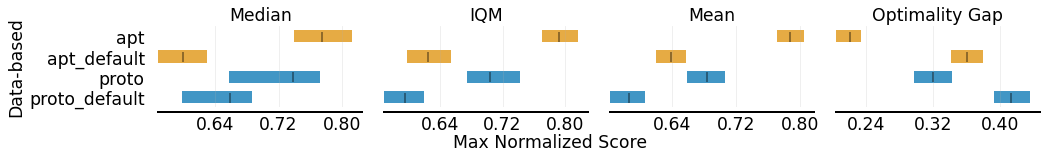}
        \label{fig:hidden_compare_data}
    }
    \hfill
    \subfloat{
        \includegraphics[width=0.98\textwidth]{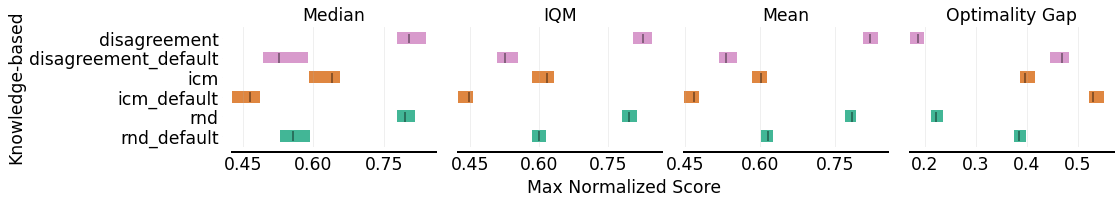}
        \label{fig:hidden_compare_knowledge}
    }
    \hfill
    \subfloat{
        \includegraphics[width=0.98\textwidth]{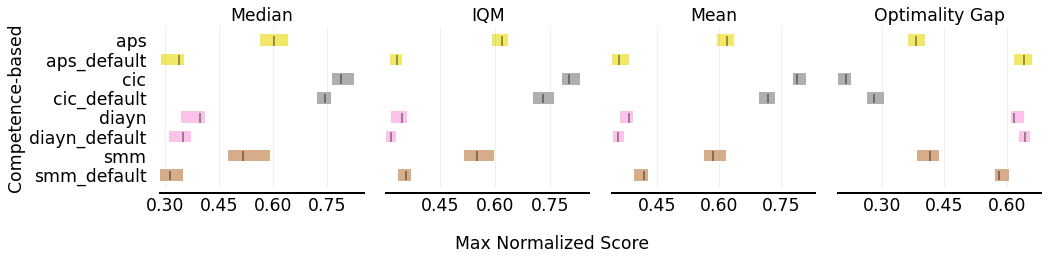}
        \label{fig:hidden_compare_competence}
    }
    \caption{\textbf{Comparing Our Baseline Results vs. Default URLB Implementaton Baseline Results.} There is an enormous performance gain in baseline algorithms using smaller latent dimension (256) compared to the default latent dimension from default URLB implementation (1024). Specifically, IQM increased by 25\%, and Optimality Gap decreased by 55\% on average.}
    \label{fig:hidden_compare}
\end{figure*}

\textbf{Investigating the effects of model capacity.}
Lastly, the advantages of our method might initially appear to be due to the increased flexibility provided by the additional parameters. However, to determine whether the improvements are actually attributable to the effectiveness of our approach rather than just the extra parameters, we expanded the baseline model's latent dimension from 256 to 1024. This step was taken to ensure that the gains observed are indeed a result of the strategic refinements of our method. Figure \ref{fig:hidden_compare} showcases a comparative analysis between baseline methods, utilizing both the original and the augmented latent dimensions. Contrary to what one might expect, the model with the \textbf{smaller} latent dimension of 256 not only registered a 25\% \textbf{increase} in the Inter-Quartile Mean (IQM) but also experienced a significant \textbf{decrease} in the Optimality Gap by 55\%. These findings challenge the prevailing assumption that larger model capacity necessarily translates to enhanced performance. Furthermore, they highlight the effectiveness of our method, which seems to more efficiently harness the additional parameters to improve the learning capabilities of URL methods. It's noteworthy that the primary results of Self-Reference (Figure \ref{fig:main_results}) were reported using the more effective hyperparameters (latent dim. = 256), underscoring the genuine impact of our approach.

\section{Conclusion and Future Work}
In this work, we presented Self-Reference, an add-on module that can enhance the effectiveness and sample efficiency of existing unsupervised reinforcement learning algorithms. We demonstrated its efficacy during pretraining and pretrain-then-finetune in more efficiently exploring the environment by mitigating the issue of nonstationarity. Moreover, we showed that our method could mitigate unlearning of useful PT behaviors by explicitly presenting the agent with its past behaviors. Finally, we evaluated our approach on the Unsupervised Reinforcement Learning Benchmark and obtained SOTA results for model-free methods. SR improved existing methods by 5\% IQM and decreased the Optimality Gap by 11\% on average.

\textbf{Limitations and Future Work.} In our work, we only explored retrieving references using a single query, therefore a natural way to enhance our method is to provide more than one query, which allows the agent to access multi-modal distributed information. Additionally, we used the state space as the key and query space for querying. This query space works well when we have the actual state of the world, but it could be inefficient if the environment is observational, e.g., images. Querying in a compact and meaningful space should significantly extend our method under these circumstances. Furthermore, we only evaluated SR with model-free URL methods, and applying SR to model-based methods could be valuable for future work. Additionally, since SR requires a retrieval and aggregation step during PT-FT stages, there is an additional computation cost during training. Works can explore retrieving not at every step to reduce training computation. Lastly, another way to extend our approach is to augment the buffer of references with expert demonstrations or any other information that could be useful in helping the agent learn about the environment and task better. Despite some progress in the field, inherent problems in the URL paradigm still need to be addressed. We believe that coherent solutions are more effective in continuing to advance this area of research.

\section*{Acknowledgement}
This work is supported in part by the National Key R\&D Program of China under Grant 2022ZD0114903, the National Natural Science Foundation of China under Grants 62022048 and 62276150,  Guoqiang Institute of Tsinghua University and Beijing Academy of Artificial Intelligence. We used AI-powered applications (i.e., https://chat.openai.com/) to polish our work at the sentence-level.


\appendix
\appendices

\section{Illustrative Example - Moving Reward Modeling}
\label{sec:motivation_pretrain}
The unsupervised reinforcement learning paradigm has two separate phases \cite{laskin2021urlb}. Current URL methods focus on exploring the environment by providing intrinsic rewards during pretraining. As the agent progressively explores the environment, the intrinsic reward of each state change according to a dynamically changing history of transitions gathered during training - and such results in a nonstationary MDP if the agent does not take this information into account \cite{sekar2020planning,jiang2022general,schafer2022decoupled}. An intuitive example of this dynamically changing reward function is the count-based exploration method where the intrinsic reward is inversely proportional to the number of visits for that state, encouraging the agent to go to less visited states by decreasing the reward of frequently visited ones \cite{bellemare2016unifying}.

Suppose we express the history of transitions obtained during training as $\Lambda_g=\{\mathbf{s}^{\pi_0}_{0},\mathbf{a}^{\pi_0}_{0},...,\mathbf{s}^{\pi_0}_H,\mathbf{a}^{\pi_0}_H,...,\mathbf{s}^{\pi_g}_0,\mathbf{a}^{\pi_g}_0,...\mathbf{s}^{\pi_g}_H,\mathbf{a}^{\pi_g}_H\},$ where $\pi_g$ indicates the changing training policy indexed by episode $g$ and $H$ is the finite number of decision epochs in an episode. We can express the PT reward as $r(\mathbf{s}, \mathbf{a};\Lambda_g)$, which implicitly depends on this history of transitions produced during pretraining. Current methods largely ignore the dependence on the history $\Lambda$ and train agents with only states and actions. Even though these methods obtain satisfactory empirical results, we posit that URL methods could be further improved if we address this issue.

Equipped with this hypothesis, we design a multi-armed bandit (MAB) experiment to verify if augmenting an agent with $\Lambda_g$ or the summary statistics of $\Lambda_g$ helps in efficiency in an environment similar to the count-based exploration setting, which could be seen as a simplified version of MDP agents will encounter during the URL pretraining phase. The multi-armed bandit environment contains $K=10$ arms, and the reward function of each of the arms is $r^k=-n^k+\mathcal{N}(0,10)$, where $n^k$ is the number of times the $k$th arm has been pulled plus a Gaussian noise with $\sigma=10$. It is obvious that the reward function is nonstationary if the agent does not utilize the summary statistics of the history of interactions. Three agents were used to interface with this environment: 1) A random agent that outputs uniformly random actions; 2) A model-based epsilon-greedy ($\epsilon=0.1$) agent that uses an exponential average to estimate the reward of each arm, and its estimates are updated using $\hat{r}^k_{t+1} = \alpha \hat{r}^k_{t} + (1-\alpha) r_t$, if the $k$th arm is pulled at time $t$, while other arms' reward estimates are not updated; 3) A model-based epsilon-greedy ($\epsilon=0.1$) agent that uses the counts' information to regress the reward. Specifically, we use least squares regression to estimate the reward, $\hat{r}^k=wn^k+b$, where $w$ and $b$ are learnable parameters, refitted at every decision epoch.

\begin{figure}[ht]
    \begin{center}
    \centerline{\includegraphics[width=\columnwidth]{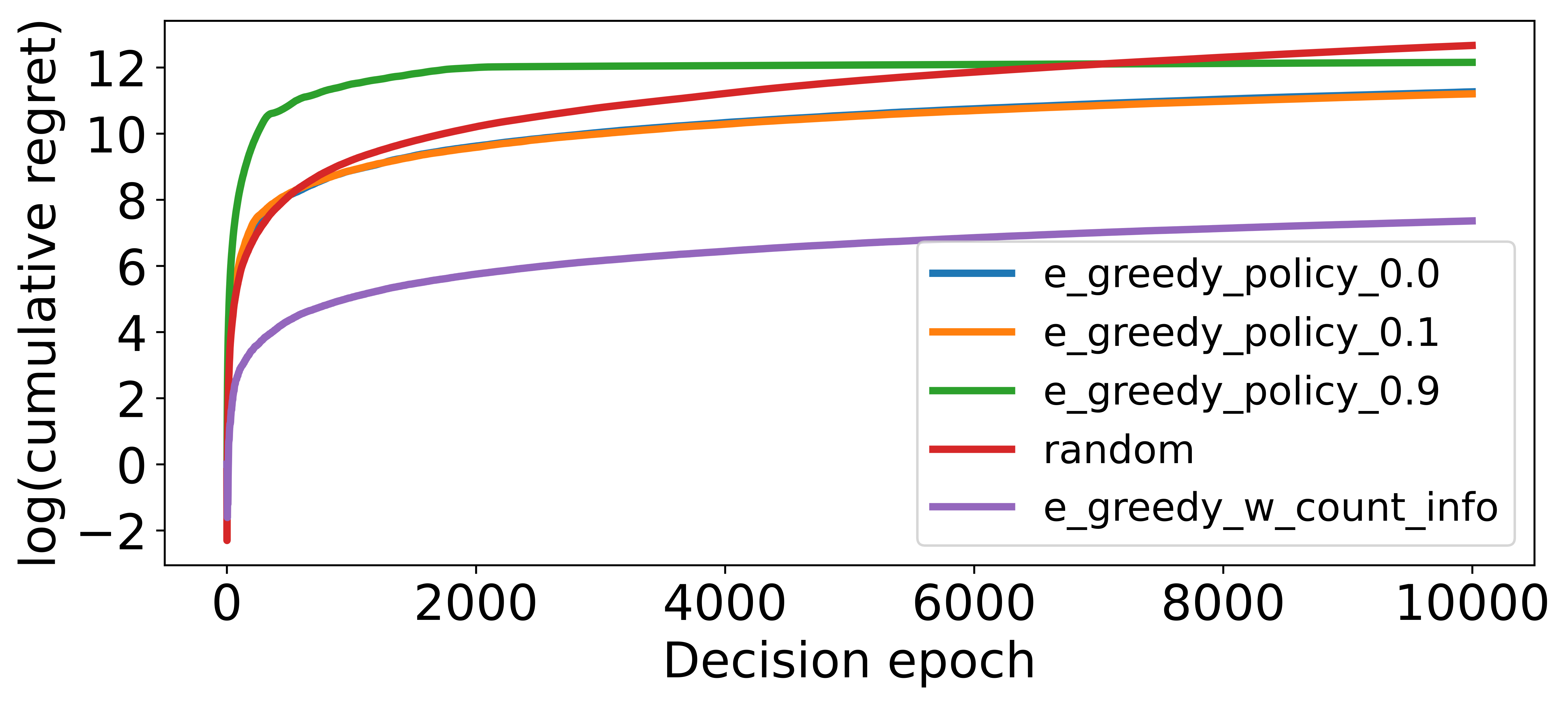}}
    \caption{\textbf{Cumulative Regret Graph for Counts Multi-armed Bandit Environment.} Plot of average regret over ten runs. We compare an agent with counts information against a random agent and exponential average agents with different $\alpha$ parameters ($\alpha\in\{0.0,0.1,0.9\}$). The agent using the historic information achieves lower regret (purple).}
    \label{mab-regret}
    \end{center}
\end{figure}

We use regret \cite{lattimore2020bandit} as the metric to look at each policy's efficiency in learning about the MDP and plot the average results over ten runs in Figure \ref{mab-regret}. We define regret as
\begin{align}
    \mathcal{R}_T=\mathbb{E}\bigg[\sum^T_{t=1}\mu^*(\nu(\Lambda_t))-X_t\bigg],
    \label{eq:regret}
\end{align}
where $\mu^*$ is the maximum mean reward based on arm $\nu$ and history of interactions $\Lambda_t$, $X_t$ is the reward obtained in round $t$. To not much surprise, the agent with the summary statistics learns about its environment more efficiently since it makes the reward function stationary. In contrast, the exponential average agent tries to "chase" the changing reward values but can never catch up. Therefore, we posit a method that could alleviate the nonstationarity reward issue by utilizing the history of transitions should aid the agent in exploring the MDP more efficiently during pretraining.

\section{Potential Negative Societal Impacts}
We operate under the subdomain of unsupervised reinforcement learning. Since the pretraining phase of the URL paradigm consists of pure exploration, the agent might be unaware of potential dangers in its environment and might cause self-harm or harm other agents that coexist with the learning agent. We believe that exploration with constraints is a possible solution to this problem. Furthermore, retrieving from past agent data could become proprietary over time and, if left unattended, could breach privacy acts. To alleviate this issue, dedicated data-cleaning infrastructure should be implemented for real-world deployment.

\section{Benchmark Details}
\label{sec:benchmark}
\subsection{Task Description}
There are three domains in the URL Benchmark, each containing four different tasks. After pretraining our agents in each domain without extrinsic reward, we finetune our agents in the four corresponding downstream tasks to evaluate URL algorithms. We briefly introduce each domain below:
\begin{itemize}
    \item \textit{Walker:} A biped constrained to a 2D vertical plane is expected to perform four specified tasks: stand, walk, flip, and run. In pretraining, Walker requires the unsupervised agent to learn basic skills like balancing and stretching to efficiently adapt to downstream tasks in finetuning.
    \item \textit{Quadruped:} A quadruped in a 3D space is expected to perform four tasks like Walker: stand, walk, flip, and run. Quadruped also needs the agent to learn basic skills in pretraining and fast adapt in finetuning, but the dimensions of state and action space are much higher than that of Walker, making learning more challenging.
    \item \textit{Jaco-Arm:} A 6-DOF robotic arm with a three-finger gripper attempts four simple manipulation tasks: reach top-left, reach top-right, reach bottom-left, reach bottom-right.
\end{itemize}
We refer readers to the original URL Benchmark paper \cite{laskin2021urlb} for a more detailed benchmark description.

\subsection{Backbone RL Algorithm - DDPG}
Since DDPG \cite{lillicrap2015continuous} is proven to be a state-of-the-art off-policy optimization algorithm and performs well in tasks from DeepMind Control Suite \cite{tassa2018deepmind}, it is chosen as the backbone algorithm of URL methods. DDPG is an off-policy RL algorithm for continuous control tasks using the Actor-Critic architecture. The critic $Q_\phi$ minimizes the Bellman error
\begin{align}
    \mathcal{L}_Q(\phi, \mathcal{D}) = & \mathbb{E}_{(s_t, a_t, r_t, o_{t+1})\sim \mathcal{D}}\nonumber\\
    & [(Q_\phi(s_t,a_t) - r_t - \gamma Q_{\overline\phi}(s_{t+1},\pi_{\theta}(s_{t+1})))^2],
\end{align}

where $\overline\phi$ is an exponential moving average of the critic weights, $\mathcal{D}$ donates the replay buffer\cite{lin1992self} and $\pi(\mathbf{s})=\arg\max_\mathbf{a}Q(\mathbf{s},\mathbf{a})$. The deterministic actor $\pi_\theta$ is trained to maximize the expected return on every state:
\begin{equation}
    \mathcal{L}_\pi(\theta, \mathcal{D}) = \mathbb{E}_{s_t \sim \mathcal{D}}[Q_\phi(s_t, \pi_\theta(s_t))].
\end{equation}

\subsection{Baselines}
To evaluate our method, we apply Self-Reference to the following baseline algorithms:

\textit{Knowledge-based} URL algorithms use prediction error as the intrinsic reward \cite{laskin2021urlb}. ICM \cite{pathak2017curiosity} and Disagreement \cite{pathak2019self} both use the forward model prediction error, while RND \cite{burda2018exploration} maximizes the prediction error of a random feature output by a random frozen network.

\textit{Data-based} algorithms explores the environment by maximizing entropy \cite{laskin2021urlb}. APT \cite{liu2021behavior} and ProtoRL \cite{yarats2021reinforcement} use k-NN particle estimators to achieve this goal.

\textit{Competence-based} URL methods differ from the previous methods by learning explicit skills and maximizing the mutual information between skills and observations \cite{laskin2021urlb}. DIAYN \cite{eysenbach2018diversity}, SMM \cite{lee2019efficient}, APS \cite{liu2021aps}, CIC \cite{laskin2022cic}, MOSS \cite{zhao2022mixture} are some staple competence-based methods on the URL Benchmark.

\section{Implementation Details}

\begin{algorithm}[tb]
   \caption{Unsupervised Reinforcement Learning with \textcolor{teal}{Self-Reference}}
   \label{alg:Self-Reference}
\begin{algorithmic}
   \State {\bfseries Input:} Initialized actor $\pi_\theta$, critic $Q_\phi$, and encoder $f_\xi$ networks, replay buffer $\mathcal{D}$.
   \State {\bfseries Input:} Intrinsic $r^\text{int}$ and extrinsic $r^\text{ext}$ reward functions, discount factor $\gamma$.
   \State {\bfseries Input:} Environment (env), $H$ maximum number of steps in episode, $M$ downstream tasks $T_k$, $k \in [1, . . . , M ]$.
   \State {\bfseries Input:} pretrain $N_\text{PT}$ and finetune $N_\text{FT}$ steps.
   \State {\bfseries Input:} \textcolor{teal}{Initialized query actor $\pi^\text{query}_\psi$, critic $V^\text{query}_\rho$, reference buffer $\mathcal{D}^\text{reference}\subset\mathcal{D}$, \texttt{GetTrajectories}, numbers of neighbors to retrieve $K$ and \texttt{NearestNeighborSearch}.}
   \For{$t=1$ {\bfseries to} $N_\text{PT}$}
   \State \textcolor{teal}{$\mathbf{q}_t\sim\pi^\text{query}_\psi(\mathbf{o_t})$}
   \State \textcolor{teal}{$\mathbf{I}_t\gets\texttt{NearestNeighborSearch}(\mathbf{q}_t,\mathcal{D}^\text{reference},K)$}
   \State \textcolor{teal}{$\mathbf{\mathcal{T}}_t\gets\texttt{GetTrajectories}(\mathbf{I}_t,\mathcal{D}^\text{reference})$}
   \State $a_t\gets\pi_\theta(f_\xi(\mathbf{o}_t),\textcolor{teal}{\mathbf{\mathcal{T}}_t})+\epsilon$ and $\epsilon\sim\mathcal{N}(0, \sigma^2)$
   \State $\mathbf{o}_{t+1} \sim P (\cdot|\mathbf{o}_t, \mathbf{a}_t)$
   \State $\mathcal{D} \gets \mathcal{D} \cup (\mathbf{o}_t, \mathbf{a}_t, \mathbf{o}_{t+1})$
   \If{$t \mod{H} = 0$}
   \State \textcolor{teal}{$\mathcal{D}^\text{reference} \gets \mathcal{D}^\text{reference} \cup \{\mathbf{o}_{t-H},...,\mathbf{o}_{t+1}\}$}
   \EndIf
   \State Update $\pi_\theta$, $Q_\phi$, and $f_\xi$ using minibatches from $\mathcal{D}$ and intrinsic reward $r^\text{int}$ using DDPG;
   \State \textcolor{teal}{Update $\pi^\text{query}_\psi$, $V^\text{query}_\rho$ using minibatches from $\mathcal{D}$ by getting batched trajectories and according to Eqs. \ref{eq:query-agent}.}
   \EndFor
   \For{$T_k\in[T_1,...,T_m]$}
   \State initialize $\theta \gets \theta_\text{PT},\phi \gets \phi_\text{PT}, \xi \gets \xi_\text{PT}$, reset $\mathcal{D}$.
   \State \textcolor{teal}{initialize $\psi \gets \psi_\text{PT},\rho \gets \rho_\text{PT}$, $\mathcal{D}^\text{reference}\gets\mathcal{D}^\text{reference}_\text{PT}$.}
   \For{$t=1$ {\bfseries to} $N_\text{FT}$}
   \State \textcolor{teal}{$\mathbf{q}_t\sim\pi^\text{query}_\psi(\mathbf{o_t})$}
   \State \textcolor{teal}{$\mathbf{I}_t\gets\texttt{NearestNeighborSearch}(\mathbf{q}_t,\mathcal{D}^\text{reference},K)$}
   \State \textcolor{teal}{$\mathbf{\mathcal{T}}_t\gets\texttt{GetTrajectories}(\mathbf{I}_t,\mathcal{D}^\text{reference})$}
   \State $a_t\gets\pi_\theta(f_\xi(\mathbf{o}_t),\textcolor{teal}{\mathbf{\mathcal{T}}_t})+\epsilon$ and $\epsilon\sim\mathcal{N}(0, \sigma^2)$
   \State $\mathbf{o}_{t+1} \sim P (\cdot|\mathbf{o}_t, \mathbf{a}_t)$
   \State $\mathcal{D} \gets \mathcal{D} \cup (\mathbf{o}_t, \mathbf{a}_t, r^\text{ext}_t, \mathbf{o}_{t+1})$
   \If{$t \mod{H} = 0$}
   \State \textcolor{teal}{$\mathcal{D}^\text{reference} \gets \mathcal{D}^\text{reference} \cup \{o_{t-H},...,\mathbf{o}_{t+1}\}$}
   \EndIf
   \State Update $\pi_\theta$, $Q_\phi$, and $f_\xi$ using minibatches from $\mathcal{D}$ using DDPG;
   \State \textcolor{teal}{Update $\pi^\text{query}_\psi$, $V^\text{query}_\rho$ using minibatches from $\mathcal{D}$ by getting batched trajectories and according to Eqs. \ref{eq:query-agent}.}
   \EndFor
   \State Evaluate performance of RL agent on task $T_k$
   \EndFor
\end{algorithmic}
\end{algorithm}

\subsection{Compute Resources}
Experiments were conducted on an internal cluster with 8 NVIDIA GeForce RTX 2080 Ti GPUs with AMD Ryzen™ Threadripper™ 3990X Processor.

\subsection{Baseline Hyperparameters}
\label{sec:hyperparameters}
\begin{table}[ht]
\caption{\textbf{Hyperparameters for Self-Reference and DDPG.}}
\centering
\begin{tabular}{@{}cc@{}}
\hline
DDPG hyperparameters        & Value                                                                                                                                           \\ \hline
\rowcolor{gray!25} Replay buffer capacity            & $10^6$                                                                                                                                          \\
Action repeat                     & 1                                                                                                                                               \\
\rowcolor{gray!25} Seed frames                       & 4000                                                                                                                                            \\
n-step returns                    & 3                                                                                                                                               \\
\rowcolor{gray!25} Mini-batch size                   & 512                                                                                                                                            \\
Discount ($\gamma$)               & 0.99                                                                                                                                            \\
\rowcolor{gray!25} Optimizer                         & Adam                                                                                                                                            \\
Learning rate                     & $10^{-4}$                                                                                                                                          \\
\rowcolor{gray!25} Agent update frequency            & 2                                                                                                                                               \\
Critic target EMA rate ($\tau_Q$) & 0.01                      
                                                            \\
\rowcolor{gray!25} Hidden dim.                       & 256                                                                                                                                             \\
Exploration stddev clip                &  0.3 \\
\rowcolor{gray!25} Exploration stddev value              &  0.2 \\
Number pre-training frames        & $2\times 10^6$                                                                                                                                          \\
\rowcolor{gray!25} Number fine-turning frames        & $1\times 10^5$                                                                                                                                          \\ \hline
Self-Reference hyperparameters              & Value                                                                                                                                           \\ \hline

\rowcolor{gray!25} Reference Context Window Size                         & 100000                                                                                                                                                                                                                                                                    \\
Neighbor sequence length $D$                             & 5                                                                                                                                         \\
\rowcolor{gray!25} Top $k$ neighbors                             & 10                                                                                                                                         \\
Identity loss coefficient $\lambda$      & 1.0  \\
\rowcolor{gray!25} Reference vector dim.       & 256  \\

\end{tabular}
\label{tab:hyperparam}
\end{table}

By referring to our main results in Figure \ref{fig:main_results}, it can be noticed that baseline methods' results significantly increased from the reported results from \textit{CIC} and \textit{MOSS} \cite{laskin2022cic,zhao2022mixture}. This discrepancy is because, while doing our experiments, we found that the size of the hidden dimension of the networks significantly impacted the performance of many algorithms. In particular, a smaller size of 256 was strictly better than 1024 for all existing methods except for a slight drop for MOSS. (We conjecture that since MOSS learns two modes of behavior, it may need a bigger network to encode the enlarged set of behaviors. We keep the 256 dimension as it drastically improved performance for all other methods.) Similar results were reported in the on-policy RL setting where smaller networks outperform bigger networks \cite{andrychowicz2020matters}. In light of this finding, we believe that the field should be encouraged to take a closer look at hyperparameters alongside designing new algorithms in the future. Overall, to have a fair comparison, we changed the networks' default hidden size to 256 for all the baselines and our method. We also reduced the batch size to 512 from 1024 and kept all the other hyperparameters identical to their original implementations.

\subsection{Self-Reference Hyperparameters}
We implement the query module's PPO component using the code and hyperparameters from CleanRL \cite{huang2022cleanrl}. For competence-based methods, the state input to the actor and critic of the query module contains the concatenation of the state and skill.

Since we interact with fixed episode lengths, if we query a neighbor that is within $D$ steps away from ending an episode, we retrieve the last $D$ steps of the episode as that particular reference trajectory.

During finetuning, we noticed that freezing the weights of the critic's module that combines the retrieved transitions into the reference vector was beneficial; therefore adopted this during FT. We hypothesize that the pretraining task correlates with the finetuning task and that updating fewer parameters makes learning dynamics more stable. Additionally, since the finetuning steps are constrained, and a drastic shift in the distribution of state visitation is needed, we relax the constraint on the query module and use it to form a query at every step (instead of using the current state as the query half of the time during PT), therefore training the query module agent at every episode during finetuning.

\subsection{\texttt{Faiss}}
\label{sec:faiss}
Because of its GPU support, we used \texttt{Faiss} \cite{johnson2019billion} as the default k-NN search library. By using the \texttt{GpuIndexFlatIP} option and by normalizing the inputs, we search for the most similar neighbors based on the cosine similarity metric,
\begin{align}
\cos\operatorname{sim}(A,B)=\frac{A\cdot B}{||A||~||B||}.
\end{align}
We posit a more dynamic metrics like learnable weights for each feature dimension could be an exciting way to extend our work for more fine-grained situation-dependent searches.

\bibliographystyle{IEEEtran}
\bibliography{main}

\begin{thebibliography}{10}
\providecommand{\url}[1]{#1}
\csname url@samestyle\endcsname
\providecommand{\newblock}{\relax}
\providecommand{\bibinfo}[2]{#2}
\providecommand{\BIBentrySTDinterwordspacing}{\spaceskip=0pt\relax}
\providecommand{\BIBentryALTinterwordstretchfactor}{4}
\providecommand{\BIBentryALTinterwordspacing}{\spaceskip=\fontdimen2\font plus
\BIBentryALTinterwordstretchfactor\fontdimen3\font minus
  \fontdimen4\font\relax}
\providecommand{\BIBforeignlanguage}[2]{{%
\expandafter\ifx\csname l@#1\endcsname\relax
\typeout{** WARNING: IEEEtran.bst: No hyphenation pattern has been}%
\typeout{** loaded for the language `#1'. Using the pattern for}%
\typeout{** the default language instead.}%
\else
\language=\csname l@#1\endcsname
\fi
#2}}
\providecommand{\BIBdecl}{\relax}
\BIBdecl

\bibitem{borgeaud2021improving}
S.~Borgeaud, A.~Mensch, J.~Hoffmann, T.~Cai, E.~Rutherford, K.~Millican, G.~B.
  Van Den~Driessche, J.-B. Lespiau, B.~Damoc, A.~Clark, D.~De~Las~Casas,
  A.~Guy, J.~Menick, R.~Ring, T.~Hennigan, S.~Huang, L.~Maggiore, C.~Jones,
  A.~Cassirer, A.~Brock, M.~Paganini, G.~Irving, O.~Vinyals, S.~Osindero,
  K.~Simonyan, J.~Rae, E.~Elsen, and L.~Sifre.
\newblock Improving language models by retrieving from trillions of tokens.
\emph{ICML}.\hskip 1em plus
  0.5em minus 0.4em\relax PMLR, 2021.

\bibitem{khandelwal2019generalization}
U.~Khandelwal, O.~Levy, D.~Jurafsky, L.~Zettlemoyer, and M.~Lewis,
  ``Generalization through memorization: Nearest neighbor language models,'' in
  \emph{ICLR}, 2019.

\bibitem{chen2022re}
W.~Chen, H.~Hu, C.~Saharia, and W.~W. Cohen, ``Re-imagen: Retrieval-augmented
  text-to-image generator,'' \emph{arXiv preprint arXiv:2209.14491}, 2022.

\bibitem{guu2020retrieval}
K.~Guu, K.~Lee, Z.~Tung, P.~Pasupat, and M.~Chang, ``Retrieval augmented
  language model pre-training,'' in \emph{ICML}.\hskip 1em plus 0.5em minus
  0.4em\relax PMLR, 2020.

\bibitem{jing2022retrieval}
B.~Jing, S.~Zhang, Y.~Zhu, B.~Peng, K.~Guan, A.~Margenot, and H.~Tong,
  ``Retrieval based time series forecasting,'' \emph{arXiv preprint
  arXiv:2209.13525}, 2022.

\bibitem{campos2021beyond}
V.~Campos, P.~Sprechmann, S.~S. Hansen, A.~Barreto, S.~Kapturowski,
  A.~Vitvitskyi, A.~P. Badia, and C.~Blundell, ``Beyond fine-tuning:
  Transferring behavior in reinforcement learning,'' in \emph{ICML 2021
  Workshop on Unsupervised Reinforcement Learning}, 2021.

\bibitem{sutton2018reinforcement}
R.~S. Sutton and A.~G. Barto, \emph{Reinforcement learning: An
  introduction}.\hskip 1em plus 0.5em minus 0.4em\relax MIT press, 2018.

\bibitem{schulman2017proximal}
J.~Schulman, F.~Wolski, P.~Dhariwal, A.~Radford, and O.~Klimov, ``Proximal
  policy optimization algorithms,'' \emph{arXiv preprint arXiv:1707.06347},
  2017.

\bibitem{laskin2021urlb}
M.~Laskin, D.~Yarats, H.~Liu, K.~Lee, A.~Zhan, K.~Lu, C.~Cang, L.~Pinto, and
  P.~Abbeel, ``URLB: Unsupervised reinforcement learning benchmark,'' in
  \emph{NeurIPS Datasets and Benchmarks Track (Round 2)}, 2021.

\bibitem{andrychowicz2020matters}
M.~Andrychowicz, A.~Raichuk, P.~Sta{\'n}czyk, M.~Orsini, S.~Girgin,
  R.~Marinier, L.~Hussenot, M.~Geist, O.~Pietquin, M.~Michalski \emph{et~al.},
  ``What matters for on-policy deep actor-critic methods? a large-scale
  study,'' in \emph{ICLR}, 2020.

\bibitem{laskin2022cic}
M.~Laskin, H.~Liu, X.~B. Peng, D.~Yarats, A.~Rajeswaran, and P.~Abbeel, ``CIC:
  Contrastive intrinsic control for unsupervised skill discovery,'' in
  \emph{NeurIPS}, 2022.

\bibitem{zhao2022mixture}
A.~Zhao, M.~G. Lin, Y.~Li, Y.-J. Liu, and G.~Huang, ``A mixture of surprises
  for unsupervised reinforcement learning,'' in \emph{NeurIPS}, 2022.

\bibitem{pathak2017curiosity}
D.~Pathak, P.~Agrawal, A.~A. Efros, and T.~Darrell, ``Curiosity-driven
  exploration by self-supervised prediction,'' in \emph{ICML}.\hskip 1em plus
  0.5em minus 0.4em\relax PMLR, 2017.

\bibitem{liu2021behavior}
H.~Liu and P.~Abbeel, ``Behavior from the void: Unsupervised active
  pre-training,'' in \emph{NeurIPS}, 2021.

\bibitem{burda2018exploration}
Y.~Burda, H.~Edwards, A.~Storkey, and O.~Klimov, ``Exploration by random
  network distillation,'' in \emph{ICLR}, 2018.
  
\bibitem{liu2021aps}
H.~Liu and P.~Abbeel, ``APS: Active pretraining with successor features,'' in
  \emph{ICML}.\hskip 1em plus 0.5em minus 0.4em\relax PMLR, 2021.

\bibitem{pathak2019self}
D.~Pathak, D.~Gandhi, and A.~Gupta, ``Self-supervised exploration via disagreement,'' in \emph{ICML}.\hskip 1em plus 0.5em minus 0.4em\relax PMLR, 2019.

\bibitem{eysenbach2018diversity}
B.~Eysenbach, A.~Gupta, J.~Ibarz, and S.~Levine, ``Diversity is all you need:
  Learning skills without a reward function,'' in \emph{ICLR}, 2018.

\bibitem{yarats2021reinforcement}
D.~Yarats, R.~Fergus, A.~Lazaric, and L.~Pinto, ``Reinforcement learning with
  prototypical representations,'' in \emph{ICML}.\hskip 1em plus 0.5em minus
  0.4em\relax PMLR, 2021.

\bibitem{lee2019efficient}
L.~Lee, B.~Eysenbach, E.~Parisotto, E.~Xing, S.~Levine, and R.~Salakhutdinov,
  ``Efficient exploration via state marginal matching,'' \emph{arXiv preprint
  arXiv:1906.05274}, 2019.

\bibitem{johnson2019billion}
J.~Johnson, M.~Douze, and H.~J{\'e}gou, ``Billion-scale similarity search with
  {GPUs},'' \emph{IEEE Transactions on Big Data}, 2019.

\bibitem{agarwal2021deep}
R.~Agarwal, M.~Schwarzer, P.~S. Castro, A.~Courville, and M.~G. Bellemare,
  ``Deep reinforcement learning at the edge of the statistical precipice,''
  \emph{NeurIPS}, 2021.

\bibitem{bellemare2016unifying}
M.~Bellemare, S.~Srinivasan, G.~Ostrovski, T.~Schaul, D.~Saxton, and R.~Munos,
  ``Unifying count-based exploration and intrinsic motivation,''
  \emph{NeurIPS}, 2016.

\bibitem{humphreys2022large}
P.~C. Humphreys, A.~Guez, O.~Tieleman, L.~Sifre, T.~Weber, and T.~P. Lillicrap,
  ``Large-scale retrieval for reinforcement learning,'' in \emph{NeurIPS},
  2022.

\bibitem{goyal2022retrieval}
A.~Goyal, A.~Friesen, A.~Banino, T.~Weber, N.~R. Ke, A.~P. Badia, A.~Guez,
  M.~Mirza, P.~C. Humphreys, K.~Konyushova \emph{et~al.}, ``Retrieval-augmented
  reinforcement learning,'' in \emph{ICML}.\hskip 1em plus 0.5em minus
  0.4em\relax PMLR, 2022.

\bibitem{vaswani2017attention}
A.~Vaswani, N.~Shazeer, N.~Parmar, J.~Uszkoreit, L.~Jones, A.~N. Gomez,
  {\L}.~Kaiser, and I.~Polosukhin, ``Attention is all you need,''
  \emph{NeurIPS}, 2017.

\bibitem{stengel1986stochastic}
R.~F. Stengel, \emph{Stochastic optimal control: theory and application}.\hskip
  1em plus 0.5em minus 0.4em\relax John Wiley \& Sons, Inc., 1986.

\bibitem{sekar2020planning}
R.~Sekar, O.~Rybkin, K.~Daniilidis, P.~Abbeel, D.~Hafner, and D.~Pathak,
  ``Planning to explore via self-supervised world models,'' in
  \emph{ICML}.\hskip 1em plus 0.5em minus 0.4em\relax PMLR, 2020.

\bibitem{jiang2022general}
M.~Jiang, T.~Rockt{\"a}schel, and E.~Grefenstette, ``General intelligence
  requires rethinking exploration,'' \emph{arXiv preprint arXiv:2211.07819},
  2022.

\bibitem{schafer2022decoupled}
L.~Sch{\"a}fer, F.~Christianos, J.~P. Hanna, and S.~V. Albrecht, ``Decoupled
  reinforcement learning to stabilise intrinsically-motivated exploration,'' in
  \emph{AAMAS}, 2022.

\bibitem{lillicrap2015continuous}
T.~P. Lillicrap, J.~J. Hunt, A.~Pritzel, N.~Heess, T.~Erez, Y.~Tassa,
  D.~Silver, and D.~Wierstra, ``Continuous control with deep reinforcement
  learning.'' in \emph{ICLR}, 2016.

\bibitem{nasiriany2022learning}
S.~Nasiriany, T.~Gao, A.~Mandlekar, and Y.~Zhu, ``Learning and retrieval from
  prior data for skill-based imitation learning,'' in \emph{CoRL}, 2022.

\bibitem{choi1999environment}
S.~Choi, D.-Y. Yeung, and N.~Zhang, ``An environment model for nonstationary
  reinforcement learning,'' \emph{NeurIPS}, 1999.

\bibitem{lattimore2020bandit}
T.~Lattimore and C.~Szepesv{\'a}ri, \emph{Bandit algorithms}.\hskip 1em plus
  0.5em minus 0.4em\relax Cambridge University Press, 2020.

\bibitem{kirk2021survey}
R.~Kirk, A.~Zhang, E.~Grefenstette, and T.~Rockt{\"a}schel, ``A survey of
  generalisation in deep reinforcement learning,'' \emph{arXiv preprint
  arXiv:2111.09794}, 2021.

\bibitem{sac}
T.~Haarnoja, A.~Zhou, P.~Abbeel, and S.~Levine, ``Soft actor-critic: Off-policy
  maximum entropy deep reinforcement learning with a stochastic actor,'' in
  \emph{ICML}.\hskip 1em plus 0.5em minus 0.4em\relax PMLR, 2018.

\bibitem{tassa2018deepmind}
Y.~Tassa, Y.~Doron, A.~Muldal, T.~Erez, Y.~Li, D.~d.~L. Casas, D.~Budden,
  A.~Abdolmaleki, J.~Merel, A.~Lefrancq \emph{et~al.}, ``Deepmind control
  suite,'' \emph{arXiv preprint arXiv:1801.00690}, 2018.

\bibitem{lin1992self}
L.-J. Lin, ``Self-improving reactive agents based on reinforcement learning,
  planning and teaching,'' \emph{Machine learning}, 1992.

\bibitem{huang2022cleanrl}
S.~Huang, R.~F.~J. Dossa, C.~Ye, J.~Braga, D.~Chakraborty, K.~Mehta, and J.~G.
  Araújo, ``Cleanrl: High-quality single-file implementations of deep
  reinforcement learning algorithms,'' \emph{JMLR}, 2022.

\bibitem{strouse2021learning}
D.~Strouse, K.~Baumli, D.~Warde-Farley, V.~Mnih, and S.~Hansen, ``Learning more
  skills through optimistic exploration,'' \emph{arXiv preprint
  arXiv:2107.14226}, 2021.

\bibitem{pmlr-v89-czarnecki19a}
\BIBentryALTinterwordspacing
W.~M. Czarnecki, R.~Pascanu, S.~Osindero, S.~Jayakumar, G.~Swirszcz, and
  M.~Jaderberg, ``Distilling policy distillation,'' in \emph{Proceedings of the
  Twenty-Second International Conference on Artificial Intelligence and
  Statistics}, ser. Proceedings of Machine Learning Research, K.~Chaudhuri and
  M.~Sugiyama, Eds., vol.~89.\hskip 1em plus 0.5em minus 0.4em\relax PMLR,
  16--18 Apr 2019, pp. 1331--1340. [Online]. Available:
  \url{https://proceedings.mlr.press/v89/czarnecki19a.html}
\BIBentrySTDinterwordspacing

\bibitem{gangwani2018policy}
T.~Gangwani and J.~Peng, ``Policy optimization by genetic distillation,'' 2018.

\bibitem{DBLP:journals/corr/abs-1806-01780}
\BIBentryALTinterwordspacing
W.~M. Czarnecki, S.~M. Jayakumar, M.~Jaderberg, L.~Hasenclever, Y.~W. Teh,
  S.~Osindero, N.~Heess, and R.~Pascanu, ``Mix{\&}match - agent curricula for
  reinforcement learning,'' \emph{CoRR}, vol. abs/1806.01780, 2018. [Online].
  Available: \url{http://arxiv.org/abs/1806.01780}
\BIBentrySTDinterwordspacing

\bibitem{DBLP:journals/corr/TehBCQKHHP17}
\BIBentryALTinterwordspacing
Y.~W. Teh, V.~Bapst, W.~M. Czarnecki, J.~Quan, J.~Kirkpatrick, R.~Hadsell,
  N.~Heess, and R.~Pascanu, ``Distral: Robust multitask reinforcement
  learning,'' \emph{CoRR}, vol. abs/1707.04175, 2017. [Online]. Available:
  \url{http://arxiv.org/abs/1707.04175}
\BIBentrySTDinterwordspacing

\bibitem{schick2023toolformer}
T.~Schick, J.~Dwivedi-Yu, R.~Dessì, R.~Raileanu, M.~Lomeli, L.~Zettlemoyer,
  N.~Cancedda, and T.~Scialom, ``Toolformer: Language models can teach
  themselves to use tools,'' 2023.

\bibitem{mialon2023augmented}
G.~Mialon, R.~Dessì, M.~Lomeli, C.~Nalmpantis, R.~Pasunuru, R.~Raileanu,
  B.~Rozière, T.~Schick, J.~Dwivedi-Yu, A.~Celikyilmaz, E.~Grave, Y.~LeCun,
  and T.~Scialom, ``Augmented language models: a survey,'' 2023.

\bibitem{qin2023tool}
Y.~Qin, S.~Hu, Y.~Lin, W.~Chen, N.~Ding, G.~Cui, Z.~Zeng, Y.~Huang, C.~Xiao,
  C.~Han, Y.~R. Fung, Y.~Su, H.~Wang, C.~Qian, R.~Tian, K.~Zhu, S.~Liang,
  X.~Shen, B.~Xu, Z.~Zhang, Y.~Ye, B.~Li, Z.~Tang, J.~Yi, Y.~Zhu, Z.~Dai,
  L.~Yan, X.~Cong, Y.~Lu, W.~Zhao, Y.~Huang, J.~Yan, X.~Han, X.~Sun, D.~Li,
  J.~Phang, C.~Yang, T.~Wu, H.~Ji, Z.~Liu, and M.~Sun, ``Tool learning with
  foundation models,'' 2023.

\bibitem{wolczyk2023role}
M.~Wolczyk, B.~Cupia{\l}, M.~Zaj{\k{a}}c, R.~Pascanu, {\L}.~Kuci{\'n}ski, and
  P.~Mi{\l}o{\'s}, ``On the role of forgetting in fine-tuning reinforcement
  learning models,'' in \emph{Workshop on Reincarnating Reinforcement Learning
  at ICLR 2023}, 2023.

\bibitem{zhao2023expel}
A.~Zhao, D.~Huang, Q.~Xu, M.~Lin, Y.-J. Liu, and G.~Huang, ``Expel: Llm agents
  are experiential learners,'' 2023.

\bibitem{castanyer2023improving}
Castanyer, Roger Creus and Romoff, Joshua and Berseth, Glen, ``Improving Intrinsic Exploration by Creating Stationary Objectives,'' 2023.
  
\bibitem{hwang2023neuro}
Hwang, Jaedong and Hong, Zhang-Wei and Chen, Eric and Boopathy, Akhilan and Agrawal, Pulkit and Fiete, Ila, ``Neuro-Inspired Fragmentation and Recall to Overcome Catastrophic Forgetting in Curiosity,'' 2023.

\end{thebibliography}

\begin{IEEEbiography}[{\includegraphics[width=1in,height=1.25in,clip,keepaspectratio]{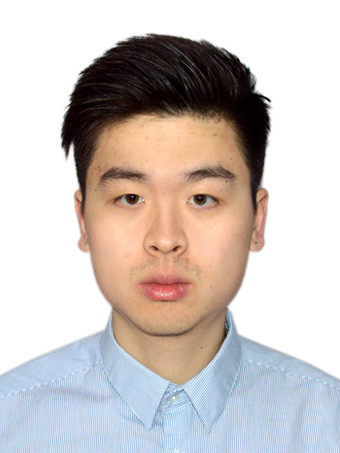}}]{Andrew Zhao} earned his Bachelor of Applied Science degree from the University of British Columbia, British Columbia, Canada, in 2017, followed by a Master of Science degree from the University of Southern California in 2020. He is currently pursuing his PhD at the Department of Automation at Tsinghua University in Beijing, China, supervised by Professor Gao Huang. His research is primarily focused on reinforcement learning, automated decision making and deep learning.
\end{IEEEbiography}

\begin{IEEEbiography}[{\includegraphics[width=1in,height=1.25in,clip,keepaspectratio]{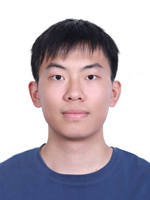}}]{Erle Zhu} is an undergraduate student of Weiyang Collge at Tsinghua University, majoring in Mathematics and Physics. His current research interest lies in reinforcement learning and natural language processing. 
\end{IEEEbiography}

\begin{IEEEbiography}[{\includegraphics[width=1in,height=1.25in,clip,keepaspectratio]{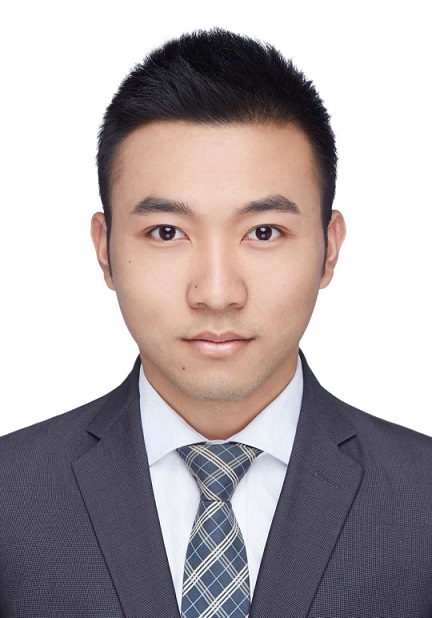}}]{Rui Lu} received his B.S. degree in Computer Science from the Institute of Interdisciplinary Information Sciences at Tsinghua University, Beijing, China, in 2021. Currently, he is pursuing his Ph.D. in the Department of Automation. Prior to this, he was a visiting student at the Berkeley Artificial Intelligence Center in 2019. His primary research interest lies in the theoretical understanding of deep learning, along with the principled design of improved learning algorithms and architectures.
\end{IEEEbiography}

\begin{IEEEbiography}[{\includegraphics[width=1in,height=1.25in,clip,keepaspectratio]{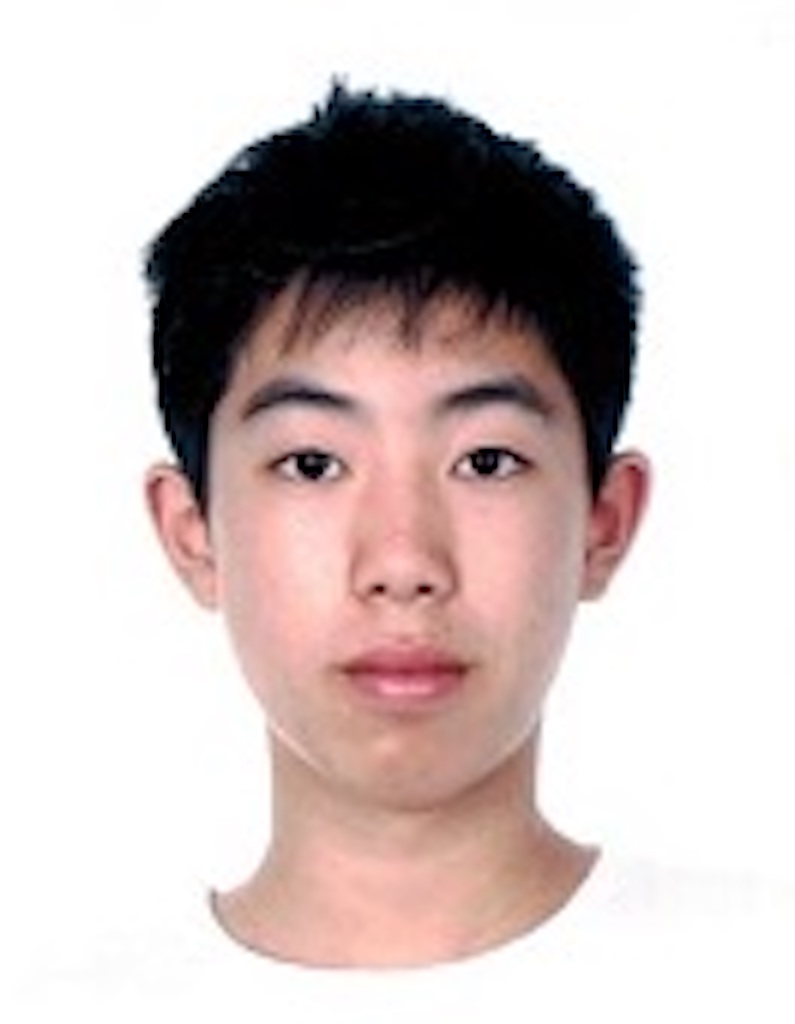}}]{Matthieu Lin} is a PhD student in the Department of Computer Science and Technology at Tsinghua University, under the supervision of Professor Yong-Jin Liu. He received his B.S.E degree in Computer Science from ESIEA Paris in 2018 and his M.S. degree in Computer Science from Tsinghua University in 2021. His research interests include reinforcement learning and computer vision.
\end{IEEEbiography}

\begin{IEEEbiography}[{\includegraphics[width=1in,height=1.25in,clip,keepaspectratio]{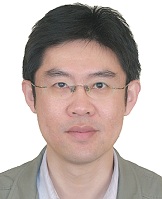}}]{Yong-Jin Liu} (Senior Member, IEEE) received the B.Eng. degree from Tianjin University, Tianjin,
China, in 1998, and the M.Phil. and Ph.D. degrees
from The Hong Kong University of Science and
Technology, Hong Kong, China, in 2000 and
2004, respectively. He is currently a Professor
with the BNRist, Department of Computer Science and Technology, Tsinghua University, Beijing,
China. His research interests include computational
geometry, computer vision, cognitive computation,
and pattern analysis. For more information, visit
https://cg.cs.tsinghua.edu.cn/people/~Yongjin/Yongjin.htm.
\end{IEEEbiography}

\begin{IEEEbiography}[{\includegraphics[width=1in,height=1.25in,clip,keepaspectratio]{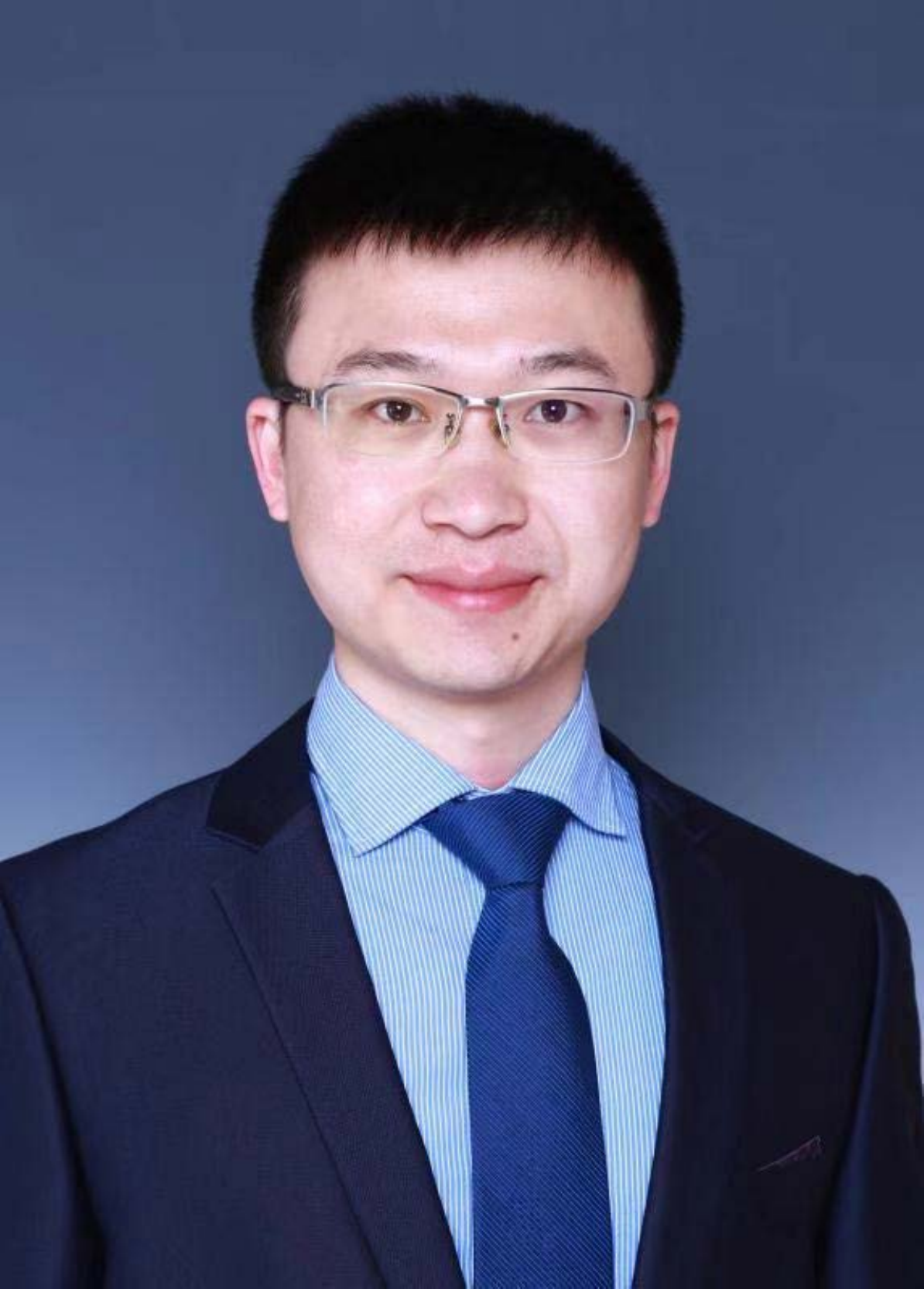}}]{Gao Huang}
received the B.S. degree from the School of Automation Science and Electrical Engineering, Beihang University, Beijing, China, in 2009 and the Ph.D. degree from the Department of Automation, Tsinghua University, Beijing, China, in 2015. From 2015 to 2018, he was a postdoctoral researcher with the Department of Computer Science, Cornell University, Ithaca, USA. He is currently an Associate Professor with the Department of Automation, Tsinghua University. His current research interests include machine learning and computer~vision.
\end{IEEEbiography}

\vfill

\end{document}